\pdfoutput=1
\documentclass[10pt,logo,copyright]{nvidiatechreport}
\usepackage[authoryear,sort&compress,round]{natbib}

\usepackage[utf8]{inputenc}
\usepackage[T1]{fontenc}

\usepackage{parskip}
\usepackage{url}
\usepackage{booktabs}
\usepackage{amsfonts}
\usepackage{nicefrac}
\usepackage[expansion=false]{microtype}
\usepackage{graphicx}
\usepackage{subcaption}
\usepackage{tabularx}
\usepackage{makecell}
\usepackage{adjustbox}
\usepackage{setspace}
\newcolumntype{M}[1]{>{\centering\arraybackslash}m{#1}}
\usepackage{float}
\usepackage{tikz}
\usetikzlibrary{positioning,shapes,arrows}
\usepackage{amsmath,amsfonts,bm,bbm,leftindex}
\usepackage{multirow}
\usepackage{comment}
\usepackage{gensymb}
\usetikzlibrary{arrows.meta,positioning,fit}
\usepackage[para]{threeparttable}
\usetikzlibrary{tikzmark}

\usepackage{wrapfig}
\usepackage{multicol}
\usepackage{amssymb}
\usepackage{nicematrix}
\usepackage{diagbox}
\usepackage{siunitx}
\usepackage{algorithm}
\usepackage{algpseudocode}

\usepackage{amsmath,amsfonts,bm}

\def\Figref#1{Figure~\ref{#1}}

\def\Secref#1{Section~\ref{#1}}

\def\eqref#1{equation~(\ref{#1})}

\def\Tableref#1{Table~\ref{#1}}

\def\1{\bm{1}}

\DeclareMathAlphabet{\mathsfit}{\encodingdefault}{\sfdefault}{m}{sl}
\SetMathAlphabet{\mathsfit}{bold}{\encodingdefault}{\sfdefault}{bx}{n}

\newcommand{\model}{GraspGen-X}
\definecolor{myGreen}{RGB}{0, 185, 0}
\definecolor{myLightBlue}{RGB}{40, 214, 253}
\definecolor{myDarkBlue}{RGB}{65,94,242}

\newcommand{\maketitlesupplementary}{%
  \clearpage
  \setcounter{section}{0}%
  \setcounter{figure}{0}%
  \setcounter{table}{0}%
  \renewcommand\thesection{\Alph{section}}%
  \renewcommand\thefigure{A\arabic{figure}}%
  \renewcommand\thetable{A\arabic{table}}%
  \section*{\model{}: Supplementary Material}%
}

\title{\model: Cross-Embodiment 6-DOF \\ Diffusion-based Grasping}

\author[1,2]{Beining Han}
\author[1]{Yu-Wei Chao}
\author[1]{Erwin Coumans}
\author[1]{Clemens Eppner}
\author[1]{Balakumar Sundaralingam}
\author[2]{Jia Deng}
\author[1]{Stan Birchfield}
\author[1]{Adithyavairavan Murali}
\affil[1]{NVIDIA}
\affil[2]{Princeton University}

\begin{abstract}
 We study cross-embodiment 6-DOF robot grasping. Unlike prior works, we require the model not only to generalize to novel objects / scenes but also to novel gripper morphologies and physical grasping processes. Our method extends diffusion model based generative 6-DOF grasping models to condition on the additional gripper's representation. We propose a swept-volume heuristic for encoding the gripper. We train our cross-embodiment model with procedural grippers and a large-scale dataset of 2 Billion grasps. In simulation experiments, our model has the best zero-shot generalization to novel real-world grippers and objects over baseline methods. Our model also serves as a good initialization for fine-tuning to adapt to novel grippers. In ablations, we demonstrate the efficiency of our sweep-volume gripper representation and our procedural gripper training dataset. Last, we show zero-shot generalization to real-world novel grippers for 6-DOF grasping, surpassing baselines in cross-embodiment generalization. 
\end{abstract}

\begin{document}

\maketitle

\vspace{-1.2em}
{\centering\fontsize{13}{16}\selectfont\href{https://github.com/NVlabs/GraspGenX}{\textcolor{nvidiagreen}{\textbf{https://github.com/NVlabs/GraspGenX}}}\par}
\vspace{0.5em}

\abscontent

\begin{figure}[h]
  \centering
  \includegraphics[width=\textwidth,keepaspectratio]{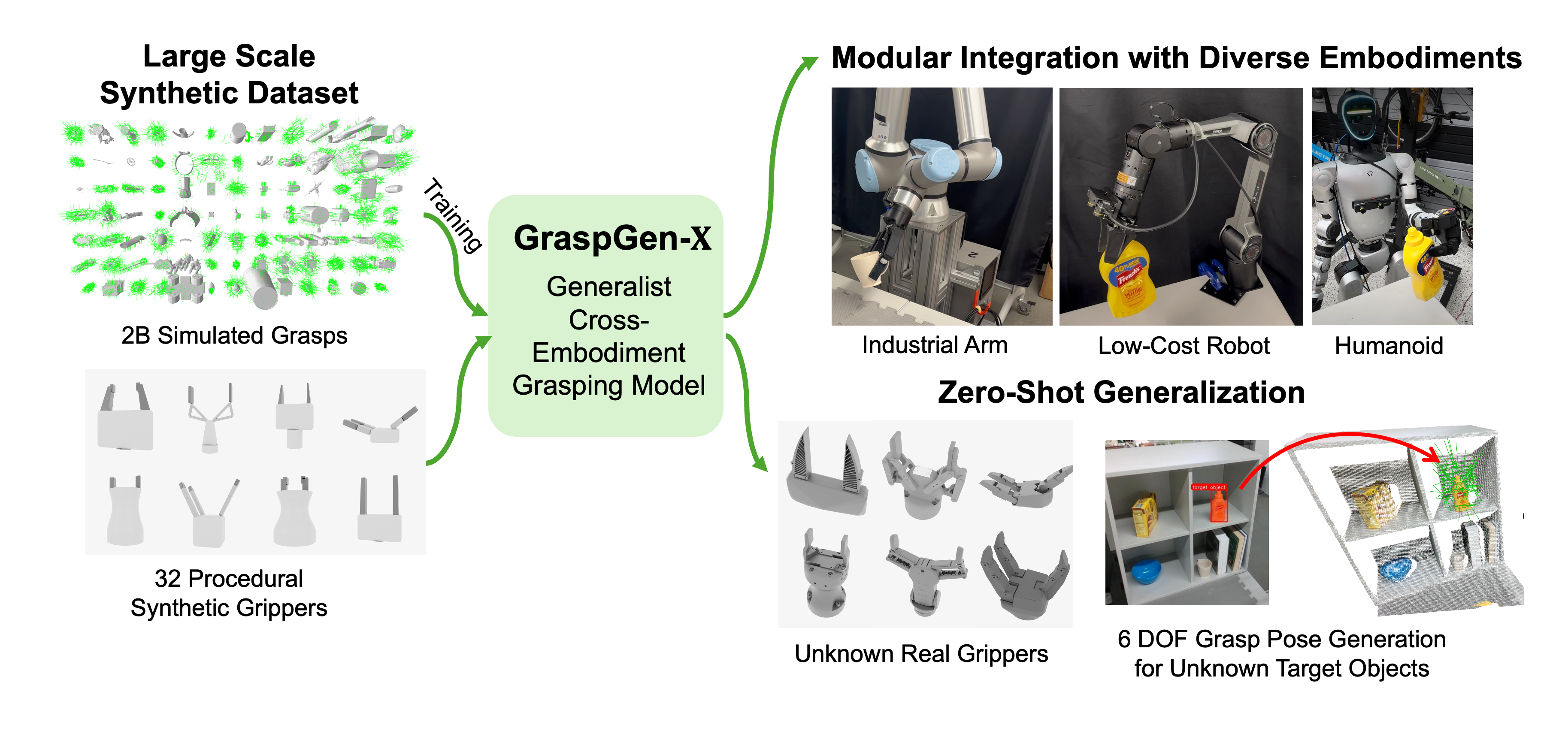}
  \vskip -0.2in
  \caption{We introduce \model{}, a cross-embodiment 6 DOF Grasping model trained with a large scale dataset of procedural grippers and 2 Billion simulated grasps. We achieve zero-shot generalization to both unknown objects as well as grippers in the real world.}
  \label{fig:coverfig}
\end{figure}

\vskip -0.1in
\section{Introduction}
\label{sec:intro}

~~~Grasping is a fundamental problem in robotics with widespread applications in industrial and household environments. In particular, the field of 6-DOF grasp generation has been significantly advanced by rapid progress in generative models \cite{murali2025graspgen, urain2022se3dif, mousavian2019-6dofgraspnet}, 3D perception \cite{wen2025foundationstereo}, physics simulation \cite{acronym2020}, and vision language models (VLM) \cite{deshpande2025graspmolmo}. It is a crucial tool-calling subroutine in recent agentic robotics systems \cite{shen2026tiptop, chen2025spacetoolstoolaugmentedspatialreasoning, li2025novaflow, fu2025capx} for prehensile manipulation.

To broaden the utility of robotic grasping in downstream applications, it is crucial to develop a generalized pick-and-place (GPnP) system. Such a system should ideally demonstrate zero-shot performance or support in-context learning -- not only for grasping novel objects across varied poses and environments, but also when deployed on a completely new robot embodiment. Current GPnP systems are typically modular, comprising interoperable components such as grasp generation \cite{fang2023anygrasp, mahler2017dexnet2, yuan2023m2t2, sundermeyer2021contact}, motion planning \cite{sundaralingam2023curobo}, occupancy mapping \cite{Millane_nvblox_ICRA2024}, and instance segmentation \cite{ravi2024sam2}, all orchestrated by a high-level planner such as a VLM \cite{deshpande2025graspmolmo, huang2024rekep, Li2025CoAVLA, Yuan2025RoboPoint, Shi2025Maestro} or task planner \cite{Wang_LearningCompositionalModels_2020, dalal2024manipgen, liu2024okrobot}. Several modules of this GPnP system have already matured and can translate zero-shot when deployed on unseen manipulators. For example, 3D perception has matured significantly, with advances in both sensing hardware and learning-based stereo models \cite{wen2023foundationpose}, and instance segmentation has benefited greatly from foundation models such as SAM2 \cite{ravi2024sam2}. Motion generation tools, such as motion planners \cite{sundaralingam2023curobo} and 3D reconstruction for occupancy mapping \cite{Millane_nvblox_ICRA2024}, are largely model-based rather than learned. As a result, adapting them to a new robot typically only requires a one-time effort to specify the robot's configuration. However, despite substantial progress in improving the generalization of 6-DOF grasp generation to unseen objects \cite{mousavian2019-6dofgraspnet, newbury2023deepgraspsurvey}, clutter \cite{Murali2020CollisionNet}, and tasks~\cite{murali2020taskgrasp}, existing methods still require retraining a new grasping model when the gripper changes. This makes grasp generation the least transferable component in cross-embodiment settings.


Recent work towards generalist robot models has explored cross-embodiment training~\cite{ONeill_OpenXEmbodiment_ICRA2024,Black_VisionLanguageActionFlowModel2024, Doshi_ScalingCrossEmbodiedLearning_CoRL2024}. For 6-DOF grasp generation, such strategy can also offer benefits. First, one can leverage a larger dataset, potentially sharing information between different embodiments to improve learned features. Second, it allows zero-shot adaptation to novel grippers and objects, reducing the need for hardware-specific training. This is especially important from the perspective of end-consumers of grasp pose generation, who may not have the time or compute resources. For example, \cite{murali2025graspgen} needs a week of an 8-GPU node to generate data and train a single-embodiment model. Many works have resorted to kinematic retargeting of grasp poses from a model trained on one embodiment and using it on a different one \cite{murali2020taskgrasp, huang2024rekep}. For example, for pinch jaw grippers, this can be achieved with a simple translation offset along the grasp approach direction. We demonstrate quantitatively that grasp pose re-targeting, while simple to implement, does not yield the best performance in large-scale simulated evaluations.

In this work, we present \model, a diffusion-based cross-embodiment 6-DOF grasp generation model that explicitly encodes a parameterized representation of the robot. Our model builds on GraspGen \cite{murali2025graspgen}, a 6-DOF grasp generation framework composed of a diffusion-based grasp pose generator and a discriminator for ranking. We extend its design by conditioning both the generator and the discriminator on the gripper's representation. Here, we propose to parameterize the gripper by its Swept Volume, defined as the region traversed by the robot fingers during its grasping motion (Sec \ref{sec:swept_volume}). With this heuristic, we demonstrate zero-shot generalization to novel grippers. Furthermore, in our training pipeline, we propose procedurally generating simulation grippers for grasp pose generation and training (Sec \ref{sec:proc_gripper}). While there are several high-quality real-world grippers that are commercially available, we find that they are not scalable to build the training dataset and the distribution is biased for learning cross-embodiment models. Our \model~is trained on a dataset of 2 Billion sampled grasps generated with the ACRONYM pipeline \cite{acronym2020} on our procedural grippers. To the best of our knowledge, this is the largest multi-embodiment dataset that has ever been used to train the grasping model.

In summary, our contributions are as follows. We present \model{}, a cross-embodiment 6-DOF grasping model and demonstrate that it has a strong zero-shot generalization to novel real-world grippers, surpassing common methods such as gripper retargeting (Sec \ref{sec:zero-shot}) and other baselines \cite{fang2023anygrasp, murali2025graspgen}. Moreover, we find that \model{} serves as a better initialization checkpoint for finetuning on a new target gripper, compared to training from scratch and finetuning from single-embodiment models (Sec \ref{sec:finetune}). We also conduct an extensive study on other common choices of gripper's representations and on comparing training with procedural grippers vs real-world grippers. We show that our Swept Volume heuristic is an efficient representation for cross-embodiment grasping, and our procedural grippers are a better training distribution. Lastly, we will open source the model, code, and dataset at \url{https://github.com/NVlabs/GraspGenX}.

\section{Related Work}
\label{sec:related_works}

\subsection{6-DOF Grasping}
~~~6-DOF grasp generation refers to the problem of predicting SE(3) grasp poses for a robot gripper given an observation of an unknown object \cite{newbury2023deepgraspsurvey}. The pose will result in the gripper stably picking up the object when the gripper closes. 
It can be applied for several downstream applications: target-driven grasping in clutter \cite{xie2024rethinking6dofgraspdetection, sundermeyer2021contact, Murali2020CollisionNet}, language-guided semantic manipulation \cite{tang2023graspgpt} and can be orchestrated by a high-level task planner \cite{dalal2024manipgen, liu2024okrobot} or VLM \cite{deshpande2025graspmolmo, huang2024rekep, Li2025CoAVLA, Yuan2025RoboPoint, Shi2025Maestro}. 6-DOF grasp generation is usually defined as a two-stage process: grasp sampling and grasp analysis \cite{newbury2023deepgraspsurvey}. Grasp sampling was initially implemented with heuristic samplers such as derivative-free optimization \cite{Levine2016LearningHandEye, mahler2017dexnet2}, analytic antipodal sampling \cite{PointNetGPD2019, tenPas2017GraspPoseDetection} or pixel-wise prediction (i.e. pixels in the image depth input are grasp pose candidates) \cite{yuan2023m2t2, sundermeyer2021contact, fang2023anygrasp}. More recently, the progress in generative models has allowed us to learn samplers in the form of autoregressive models \cite{tobin2018grasp, yuan2023m2t2}, Variational Autoencoder (VAE) \cite{mousavian2019-6dofgraspnet, Murali2020CollisionNet}, flow-matching \cite{lim2024equigraspflow} and diffusion models \cite{wu2023learning-dex-human-affordance, urain2022se3dif, lum2024get, barad2024graspldm, freiberg2024diffusionmultiembodimentgrasping, carvalho2024graspdiffusionnetworklearning}. Grasp analysis typically entails discriminator models to score sampled grasps \cite{weng2024dexdiffuser, mousavian2019-6dofgraspnet, song2024implicitgraspdiffusion, Murali2020CollisionNet}. Recent methods such as GraspGen \cite{murali2025graspgen} propose a powerful combination of diffusion-based grasp sampling and discriminators trained through an interleaved data flywheel. Our work is concerned with cross-embodiment models that can generalize not only to novel objects but also to novel grippers.


\subsection{Cross-Embodiment Learning in Robotics}
~~~Cross-embodiment learning is a challenging transfer learning problem in robotics. Prior work has proposed many methods: learning an implicit encoding of robot \cite{Chen_Murali_Gupta_HardwareConditionedPolicies_CVPR2018, Gupta_Murali_Gandhi_Pinto_RobotLearningInHomes_NeurIPS2018}, conditioning on an explicit representation of robot properties \cite{Chen_Murali_Gupta_HardwareConditionedPolicies_CVPR2018}, reward functions \cite{Zakka_XIRL_CoRL2021}, modular policies \cite{Devin_LearningModularNeuralNetworkPolicies_ICRA2017, Huang_OnePolicyToControlThemAll_ICML2020}, model-based RL \cite{Hu_KnowThyself_ICLR2022, Salhotra_LearningRobotManipulation_CoRL2023} and test-time adaptation \cite{Yu_LearningUniversalPolicy_RSS2017}. Action retargeting has been a popular approach, where actions from a source embodiment are mapped to a target embodiment, such as in tele-operation of dexterous hands \cite{Qin_AnyTeleop_2023, Handa_DexPilot_ICRA2020}. More recently, Vision-Language-Action (VLA) models such as RT-X \cite{ONeill_OpenXEmbodiment_ICRA2024} have proposed general-purpose architectures that can be trained by aggregating datasets across diverse robots \cite{Bousmalis_RoboCat_TMLR2023, Black_VisionLanguageActionFlowModel2024, ONeill_OpenXEmbodiment_ICRA2024, Doshi_ScalingCrossEmbodiedLearning_CoRL2024} and using methods like readout tokens \cite{Doshi_ScalingCrossEmbodiedLearning_CoRL2024} or masking \cite{Black_VisionLanguageActionFlowModel2024} to account for variable action space. VLAs still do not achieve strong zero-shot performance in completely new robot hardware or tasks. We hypothesize that an explicit parameterization of the embodiment, such as our Swept Volume representation in Sec~\ref{sec:swept_volume}, is necessary to achieve zero-shot performance on unknown robots in grasping.

\subsection{Multi-Embodiment Grasping}
~~~In the context of robot grasping, prior work has proposed both new datasets \cite{Casas_MultiGripperGrasp_IROS2024, Li_GenDexGrasp_ICRA2023} and algorithms \cite{Freiberg_DiffusionForMultiEmbodimentGrasping_2025, shao2020unigrasp, Attarian_GeometryMatchingMultiEmbodimentGrasping_CoRL2023, xu2021adagrasp} for multi-gripper grasping problem. \textit{Object-centric} methods have used to learn a Contact Map \cite{Li_GenDexGrasp_ICRA2023}, contact points \cite{Attarian_GeometryMatchingMultiEmbodimentGrasping_CoRL2023, shao2020unigrasp} or dense point-wise contact correspondences between the object and robot \cite{Wei_DROGrasp_ICRA2025, Fei_TROGrasp_2025}. They then use inverse kinematics (IK) to generate joint configurations and align robot contact points. These approaches assume access to a complete 3D geometry of the object, hence they do not generalize to partial point cloud observations where reliable contact points are occluded. Several \textit{gripper-aware} methods use various techniques to encode the robot: \cite{Wang_TransferringGraspingAcrossGrippers_TRO2024} uses a parameterized fingertip antipodal representation, UniGrasp \cite{shao2020unigrasp} and $\mathcal{D}(\mathcal{R},\mathcal{O})$ \cite{Wei_DROGrasp_ICRA2025} uses a conditional Variational Autoencoder (cVAE), AdaGrasp \cite{xu2021adagrasp} uses a Truncated Signed Distance Field (TSDF) representation, and \cite{Freiberg_DiffusionForMultiEmbodimentGrasping_2025} use a PointNet++ \cite{qi2017pointnet++} backbone. All such methods are trained on just a handful of real-world grippers and hence struggle to generalize to new grippers. In contrast, we train on procedural synthetic grippers and propose a novel gripper representation for cross-embodiment grasping. Ours achieves strong zero-shot generalization to unknown real grippers.  
\section{\model: Cross-Embodiment Grasping}
\label{sec:method}

~~~In the 6-DOF cross-embodiment robot grasping problem, we are given the object/scene pointcloud, the gripper's URDF, and its gripper closing motion, i.e., the joint trajectory of the gripper from fully open to fully closed. The objective is to predict SE(3) grasp poses that can successfully grasp the object. Most importantly, our aim is to generalize to both novel grippers and objects (Fig \ref{fig:coverfig}).


Our model is developed from GraspGen \cite{murali2025graspgen}, the SOTA diffusion-based grasp pose generator, which balances well between grasping accuracy and grasp pose coverage. GraspGen is composed of a diffusion model for SE(3) grasp pose generator and a discriminator to estimate the accuracy of the generated grasps. In GraspGen, the diffusion-based generator diffuses over the SE(3) space and is conditioned on the object embedding. The object embedding is encoded from object pointclouds with a PointTransformer~\cite{wu2024ptv3} or PointNet++~\cite{qi2017pointnet++}. The discriminator is trained with on-generator data of positive and negative grasp poses to predict the confidence of each grasp. It is conditioned on the same object embedding.

\begin{figure}[t]
  \centering
  \includegraphics[width=\linewidth]{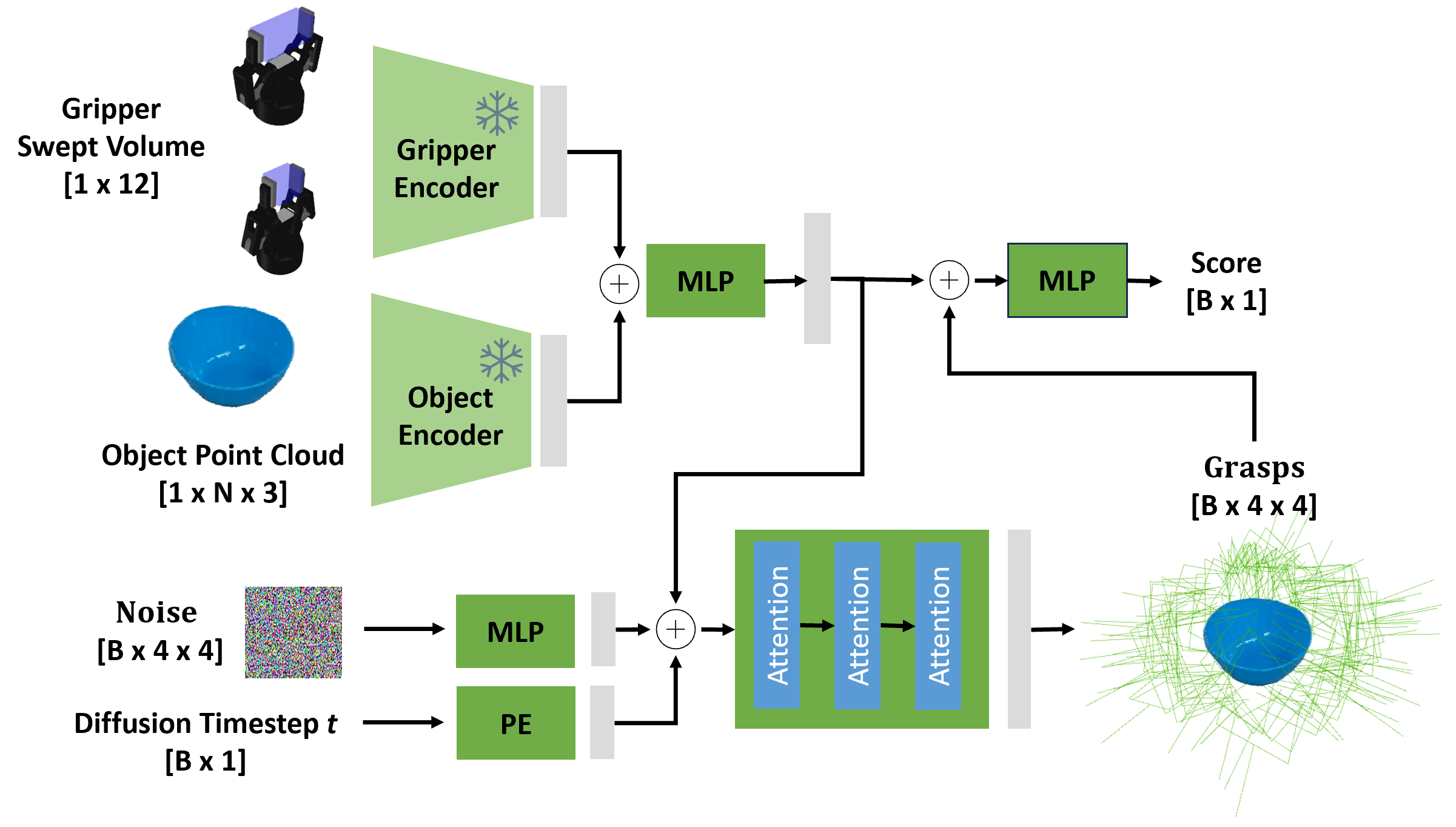}
  \vskip -0.1in
  \caption{Architecture for our \model~model. The model is developed based on GraspGen \cite{murali2025graspgen}. We additionally condition on the gripper's representation for both the generator and the discriminator. The gripper embedding is computed from the Swept Volume heuristic (Sec \ref{sec:swept_volume}).}
  \vskip -0.2in
  \label{fig:xgrasp_arch}
\end{figure}

For our cross-embodiment model, we additionally condition on the embedding of gripper. \Figref{fig:xgrasp_arch} shows the architecture of \model~, which is an extension of the GraspGen (Fig. 2 in \cite{murali2025graspgen}). The gripper embedding is encoded with a 3-layer MLP of the Swept Volume heuristic when the gripper is fully open and half open (\Secref{sec:swept_volume}).

\subsection{Gripper Encoding with Swept Volume}
\label{sec:swept_volume}

~~~Our work includes gripper categories of parallel grippers (e.g., Franka Panda Hand), 2-finger revolute grippers (e.g., Robotiq-2F85), and high-dof 3-finger grippers (e.g., Unitree G1 7-DOF Hand). To encode the gripper's morphology and closing motion into an efficient embedding for grasping models, we propose to use the Swept Volume of the gripper when it is fully open and half open. 

\begin{figure}[t]
  \centering
  \includegraphics[width=0.9\linewidth]{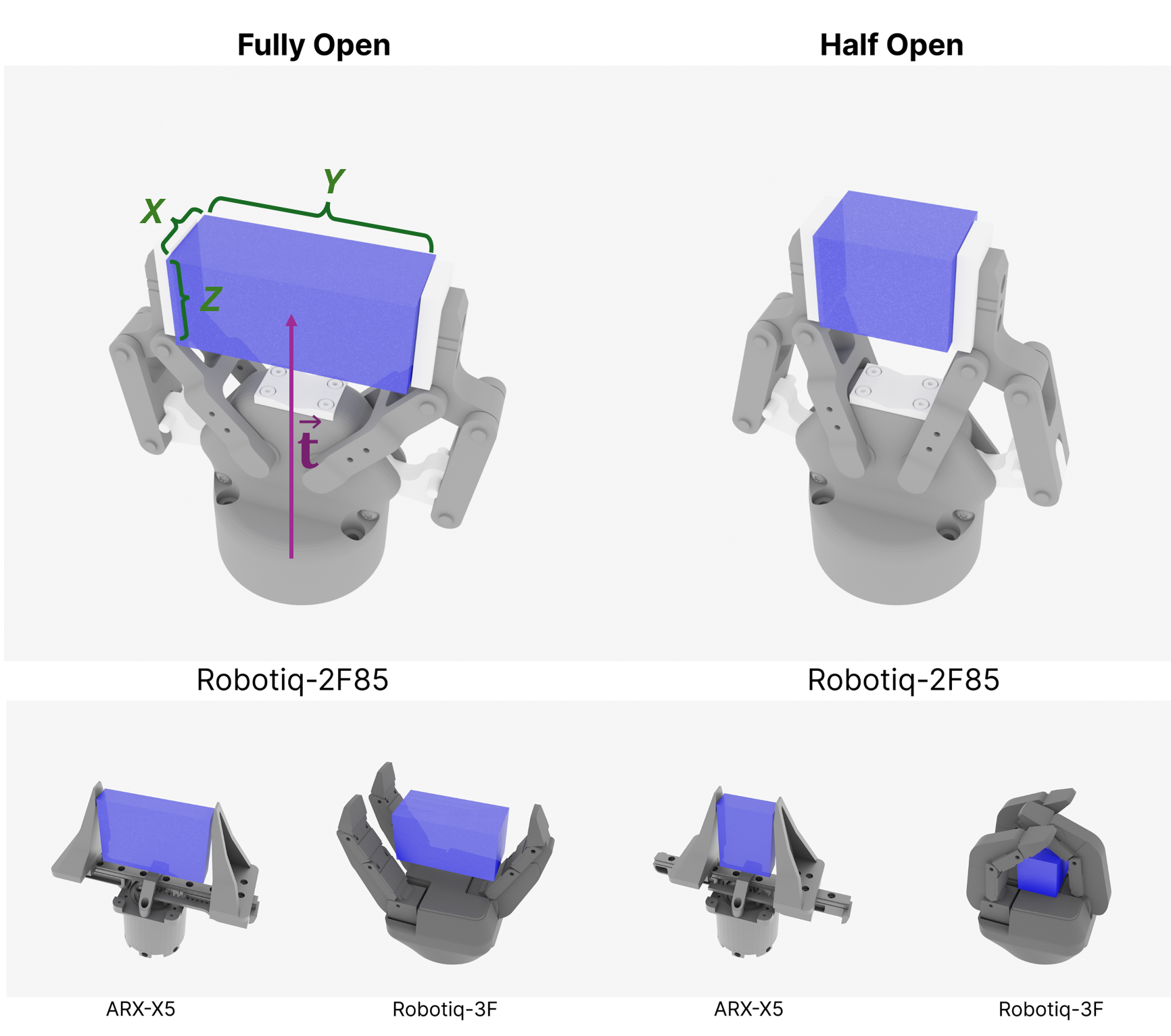}
  \caption{Illustration of the Swept Volume, i.e., cube dimensions (xyz) and cube center translation relative to gripper base ($\mathbf{t}$). We visualize the Swept Volume cube with the blue cube, when grippers are fully open (top-L) and half open (top-R). The Swept Volume varies between grippers (bottom).}
  \label{fig:swept_volume}
  \vskip -0.2in
\end{figure}

Swept Volume is a heuristic to approximate the space that gripper fingers will sweep through during the closing process. As shown in Fig \ref{fig:swept_volume}, we approximate the space with an axis-aligned cube. Thus, each Swept Volume consists of the cube dimension (3-dim) and the translation of the center from the gripper's base frame (3-dim). We use the Swept Volume (6-dim) heuristic when the hand is fully open and when it is half-way through the closing process. The gripper's representation input is a 12-dim vector in total. We encode it with a simple 3-layer MLP to the 512-dim gripper embedding to both the geneartor and the discriminator.

As illustrated in Fig \ref{fig:swept_volume}, we determine the dimension and translation of Swept Volume in the following way. For 2-finger parallel grippers (e.g., Franka Panda Hand) and revolute grippers whose fingers are always parallel during the closing process (e.g., Robotiq-2F85), we estimate a Swept Volume to cover the space between the two fingers. For grippers whose fingers are not always parallel but will rotate w.r.t the gripper base (e.g., Robotiq-3F), we approximate the space that the finger will sweep through in the follow-up process with a cube.

\subsection{Procedural Gripper Generation}
\label{sec:proc_gripper}

~~~Real-world grippers are limited in quantity and diverse in geometry and physical closing motion. In our work, we collect a total of 20 real-world grippers (\Figref{fig:gripper_train_test}, Appendix) and evenly divide them into 10 training grippers and 10 test grippers. In \Secref{sec:ablation_study}, we find that training with limited real-world grippers has relatively poor performance, due to the distribution mismatch between training/test sets and the relatively scattered data points in gripper's space. 

\begin{figure*}[t]
  \centering
  \includegraphics[width=0.9\linewidth]{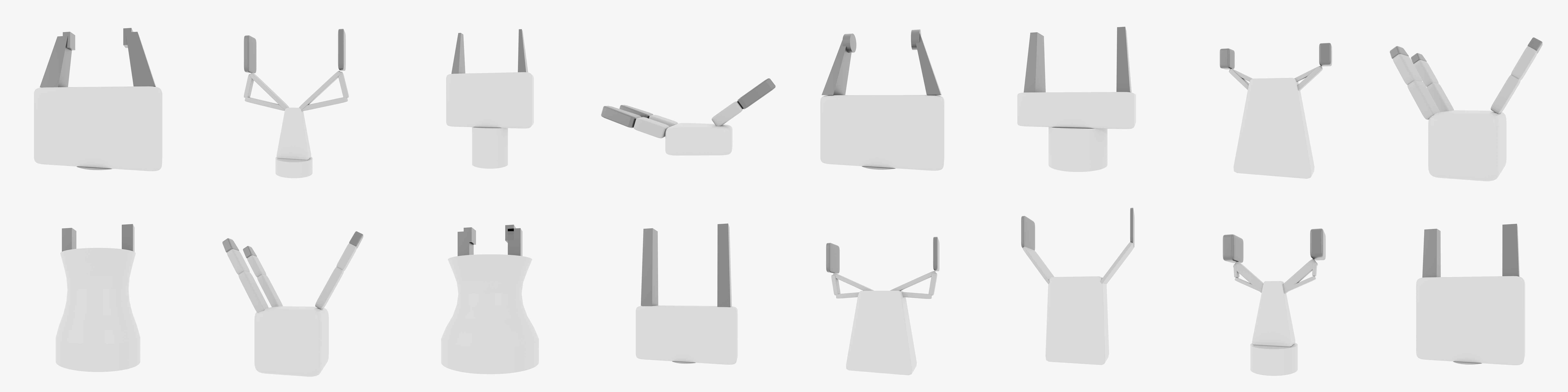}
  \caption{Examples of our procedural grippers. Each gripper is a random instance from our generator (Sec \ref{sec:proc_gripper}). Please refer to the appendix for generator details.}
  \label{fig:proc_grippers}
  \vskip -0.2in
\end{figure*}

Consequently, we propose to train the cross-embodiment model with procedural grippers randomly generated from the procedural robot gripper generator. Our procedural gripper generator is implemented with Infinigen-Sim \cite{joshi2025infinigensim}, which leverages Blender's geometry node to devise procedural mathematical rules to compose articulated objects under random configurations. Fig \ref{fig:proc_grippers} shows examples of procedural grippers randomly generated with our generator. 

We design a procedural generator class for each gripper category, i.e., parallel grippers, revolute 2-finger grippers, and 3-finger high-dof grippers. As our grippers are only used in simulation-based grasping data generation and in learning gripper's embedding, we do not need to model the fine details like screws and connectors in designing CAD models. Instead, we focus on the diversity of the overall size / morphology and on the diversity and realism of finger geometry, which will likely come into contact with the objects during the closing process. Additionally, our procedural generator also outputs the Swept Volume and other metadata needed for model training and data generation. Please refer to the appendix for more details of the generators.

\begin{figure}[t]
  \centering
  \includegraphics[width=0.9\linewidth]{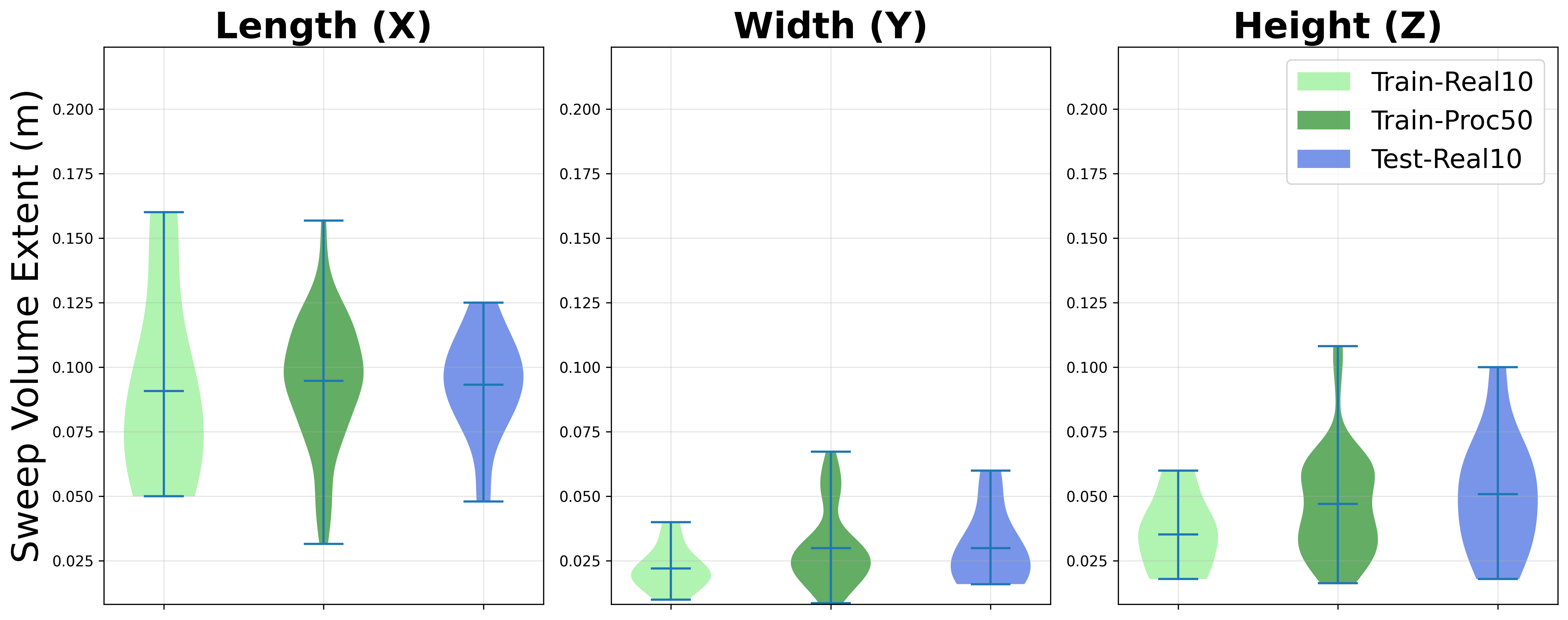}
  \caption{The distribution of Swept Volume cube dimensions along three axes, for training datasets (Train-Proc50 and Train-Real10) and a test dataset consisting of 10 novel real grippers (Test-Real10). All grippers are fully open.} 
  \label{fig:proc_dist_sv}
  \vskip -0.2in
\end{figure}

To determine the distribution of the random configurations to our generator, we leverage the set of real-world training / test grippers. We tune the randomization range so that the Swept Volume heuristics cover the real-world counterpart's distribution. Fig \ref{fig:proc_dist_sv} shows the distribution of Swept Volume dimensions of fully open grippers, comparing between 50 procedural grippers, 10 real training grippers, and 10 real test grippers. The distribution of the 10 real training grippers (Real-Train10) has some regions which are non-overlapping and out of distribution compared to the test set of grippers (Real-Test10). In contrast, our procedural gripper dataset (Proc-Train50) has a higher overlap in-distribution with the test set (Real-Test10), demonstrating the potential for superior generalization results.


\section{\model~Dataset}
\label{sec:dataset}

~~~To train the cross-embodiment model for zero-shot generalization, we have generated a large-scale grasping dataset. We follow the same antipodal sampling and simulation-based grasp labeling pipeline in ACRONYM \cite{acronym2020}, which is widely used in previous work \cite{sundermeyer2021contact, murali2025graspgen, yuan2023m2t2}. For 3-finger high-dof grippers (e.g., Unitree G1 7-DOF Hand), we use the same antipodal sampler and find that it provides a reasonable grasp distribution for 3-finger gripper grasping.

In our experiment, we use a subset of 3.5K training objects used in GraspGen \cite{murali2025graspgen} training and 453 objects as test objects. For all test objects, we ensure that each of 10 test grippers has at least 5 positive grasps to compute the coverage metric. To train the cross-embodiment model, we randomly generate 25 procedural grippers, with 10 parallel grippers, 10 2-finger revolute grippers, and 5 3-finger high-dof grippers. For each gripper and each object, we sample a maximal of 2K grasps and utilize Isaac-Sim \cite{isaacsim} to simulate the grasp outcome. In total, we have sampled and evaluated 175M grasps for generator training, which takes approximately 8.7K GPU hours. The generator is trained with 8 A100 GPUs for 780K steps with a learning rate $1e\text{-}5$, which spans 80 hours.

To generate the on-generator data for discriminator training, we generate 2K grasps for each object and each gripper, and then we evaluate grasps with the same labeling pipeline, which have taken 5.2K GPU hours. We train the discriminator with 50\% on-generator positive grasps and 50\% on-generator negative grasps. The discriminator is trained with 8 A100 GPUs at a learning rate of $1e\text{-}5$ for 300K steps, which spans 76 hours. Please refer to the appendix for details of the \model~dataset and training.

In addition to the CVPR model, we also provide our latest \model{}, which is trained on 8.5K objects, 32 procedural grippers and a total of more than 2 Billion grasps. Please refer to the Appendix for details.
\section{Simulation Experiments}
\label{sec:sim_exp}


~~For simulation-based experiments, we follow the full pointcloud experiments and evaluation metrics in GraspGen\cite{murali2025graspgen}. For test grippers and test objects, we sample 5K grasps to curate the ground-truth grasp dataset with the ACRONYM pipeline. For each test object and gripper, the generator generates 2K grasps, validated with the discriminator. We use ranked grasps to produce the precision-recall (PR) curve. We report the final AUC value of the PR curve, averaged over all test objects and test grippers.

\subsection{Zero-shot Evaluation}
\label{sec:zero-shot}

We compare \model{} with the following baselines.

 \begin{enumerate}
    \item[-] \textbf{GraspGen Direct Transfer (DTR)}: We train a GraspGen model for the Franka gripper over the same set of 3.5K training objects and with 7M sampled grasps, and directly apply the model on test grippers.
    \item[-] \textbf{GraspGen Retargeting (RTG)}: We use the GraspGen (DTR) model but retarget the predicted grasp poses. Specifically, we apply an offset to the grasp pose along the approach direction, based on the distance of the fingertip position between the test gripper and Franka. This process is commonly used for adapting a 6-DoF grasp pose to a new gripper \cite{murali2020taskgrasp, huang2024rekep, Yamada_GraspMPC_2025}. 
 \end{enumerate}

 \begin{table*}[ht]
      \centering
      \caption{Zero-shot performance on novel test grippers and novel test objects. We report the average of 4 parallel grippers, 4 revolute 2-finger grippers, 2 high-dof 3-finger grippers, and all 10 grippers. }
      \begin{tabular}{ccccc}
        \toprule
        & Parallel 2F & Revolute 2F & High-dof 3F & All \\
        \midrule
        GraspGen-DTR & 0.215 & 0.033 & 0.136 & 0.126 \\
        GraspGen-RTG & 0.365 & 0.379 & 0.503 & 0.398 \\
        \model~(\textbf{Ours}) & \textbf{0.502} & \textbf{0.413} & \textbf{0.699} & \textbf{0.506} \\
        \bottomrule
      \end{tabular}
      \label{tab:zero-shot}
\end{table*}

\Tableref{tab:zero-shot} shows the zero-shot performance of 10 test grippers by category. GraspGen-DTR completely fails to adapt to grippers that fall in a different category, e.g., revolute 2-finger grippers. GraspGen-RTG achieves a relative improvement of over 200\% compared to DTR, suggesting that the widely used heuristic is an efficient technique. However, it only considers the z-axis offset between different grippers' fingertips, and fails to consider the difference in the finger geometry and contact dynamics. Thus, there is still significant room for improvement, even in the parallel 2-finger gripper category.

In contrast, \model{} achieves the SOTA performance in all categories, further improving over GraspGen-RTG by 25\%. This suggests that it is more promising to learn an end-to-end cross-embodiment model rather than applying a simple pose correction technique to single-embodiment models. This is especially the case for high-dof 3-finger grippers where the relative improvement is nearly 40\%. The gap between different grippers comes from a mixture of morphology difference, finger geometry difference, and the difference in the physical process of hand closing.

We also evaluated \model{} on two \textit{out-of-the-distribution} 5-finger grippers, that are not involved during the training. Surprisingly, our model can still achieve a relatively high performance of 0.404 on Surge Hand and 0.363 on Inspire Hand.

\begin{figure}[t]
  \centering
  \includegraphics[width=\linewidth]{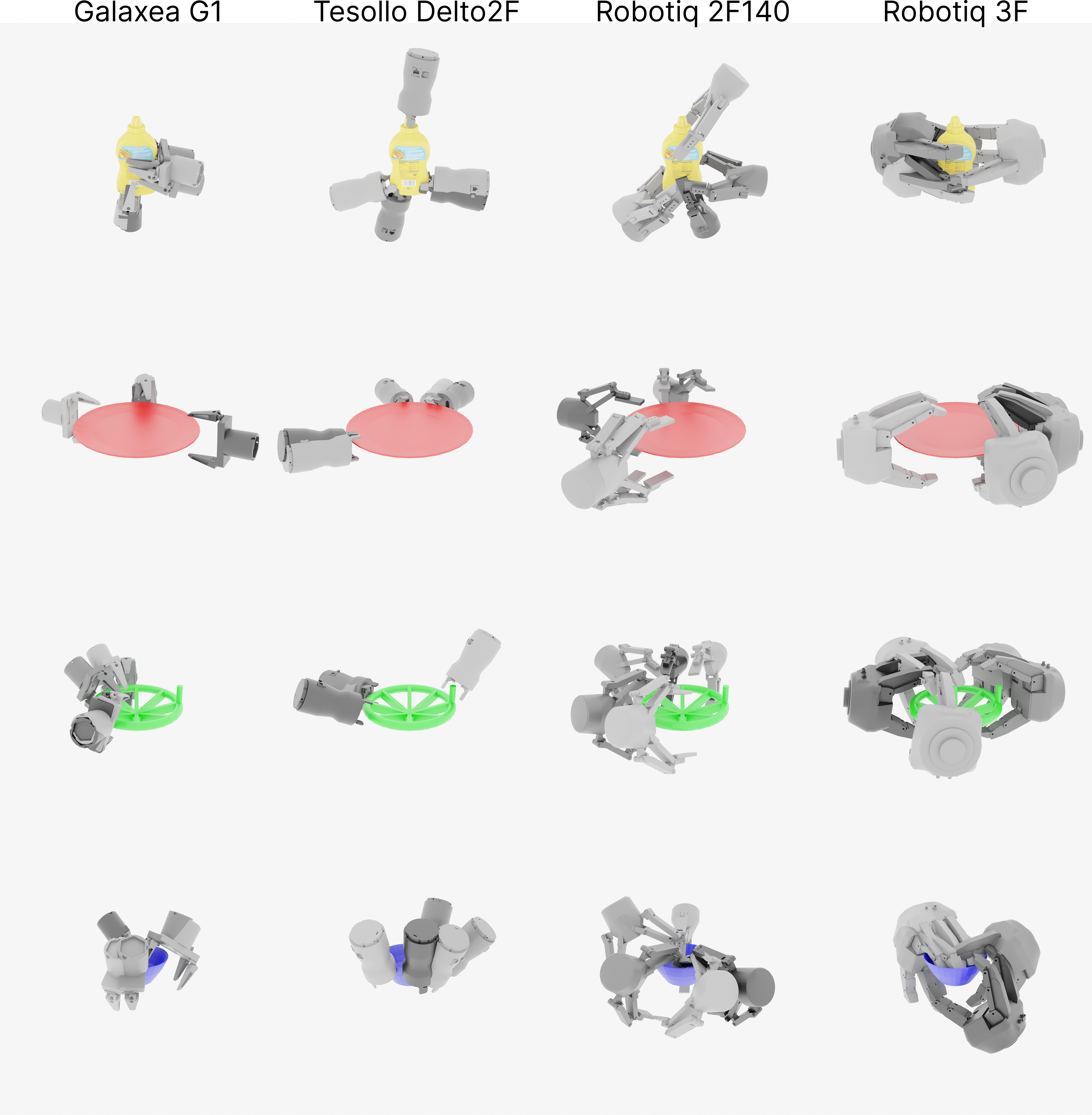}
  \caption{Visualization of generated grasp poses with \model{} on Galaxea-G1, Tesollo-Delto2F, Robotiq-2F140, Robotiq3F grippers on 4 novel objects.}
  \label{fig:graspgenx_grasps}
  \vskip -0.2in
\end{figure}

Additionally, Fig \ref{fig:graspgenx_grasps} visualizes the grasp poses predicted by ours \model{} of a novel object with novel test grippers. Please refer to the appendix for more results.

\subsection{Supervised Finetuning Adaptation}
\label{sec:finetune}

~~~We show that our \model{} is a better initialization for supervised finetuning (SFT) on novel grippers. We randomly select $1/5$ of all training objects to generate a small finetuning dataset of 140K sampled grasps following the same pipeline. For each gripper, this takes approximately 28 GPU hours of computation. We then finetune the generator on 4 A100 GPUs for 4 hours for each gripper separately. Here, we plot the learning curve of the generator's metrics on the test objects, averaged over all 10 test grippers. Similar to \cite{murali2025graspgen}, we use metrics of grasp pose translation / rotation error and the recall rate of ground truth positive grasps. Please refer to the Appendix for the description of these metrics.

\begin{figure}[t]
  \centering
  \includegraphics[width=\linewidth]{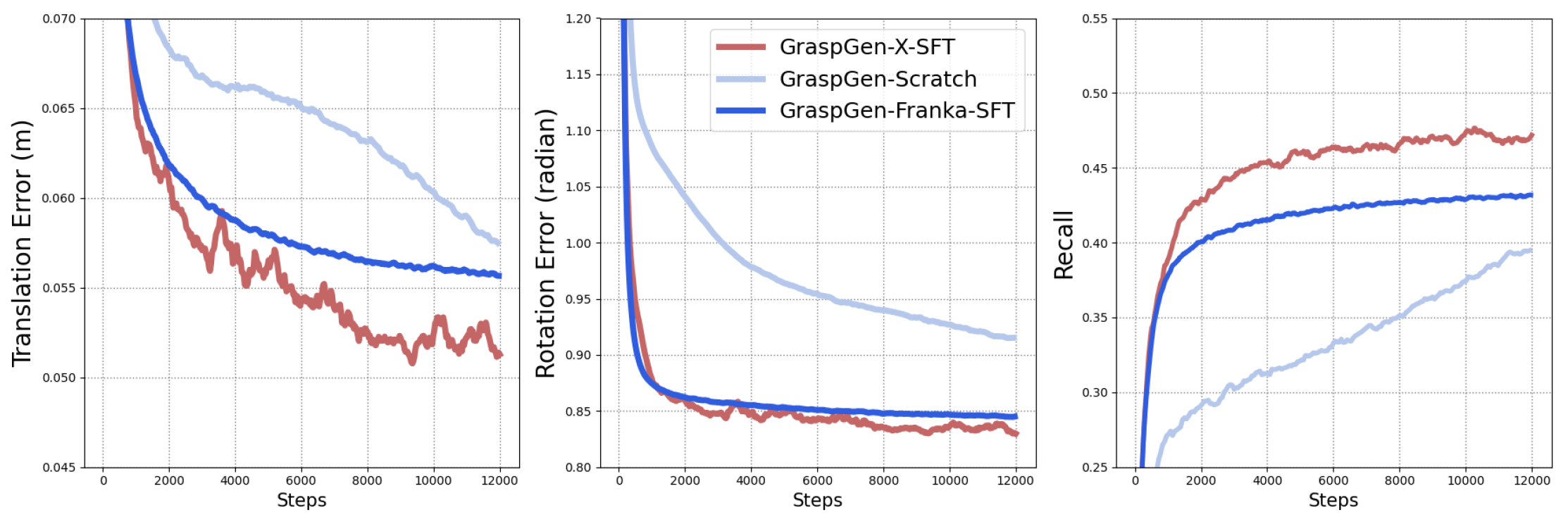}
  \caption{Learning curve of generator finetuning (Sec \ref{sec:finetune}). We plot the metric of translation error, rotation error, and recall rate of the target gripper along the training. The curve is averaged over all 10 test grippers.}
  \label{fig:finetune_curve}
  \vskip -0.1in
\end{figure}

Fig \ref{fig:finetune_curve} shows the learning curve of three metrics when training a GraspGen model from scratch (GraspGen-Scratch), finetuning the Franka Panda GraspGen in Sec \ref{sec:zero-shot} (GraspGen-Franka-SFT) and finetuning \model{} in Sec \ref{sec:method} (GraspGen-X-SFT). For GraspGen-Franka-SFT and GraspGen-X-SFT, we use a learning rate of $1e\text{-}6$, which we find to be the most efficient. For GraspGen-Scratch, we use the same $1e\text{-}5$ learning rate as in Sec \ref{sec:method}.

GraspGen-X-SFT shows the most efficient learning compared to GraspGen-Franka-SFT and GraspGen-Scratch, suggesting that the \model{} model is a better initialization when adapting to a novel gripper. We believe that the ability to cross-embodiment \model{} makes the model more adaptive to a novel gripper representation and a novel 6-dof grasping dataset.

\subsection{Gripper Encoder Comparison}
\label{sec:gripper_encoder}

~~~We compare our Swept Volume representation for the gripper encoding (Fig \ref{fig:swept_volume}) with the following baseline encodings used in prior work.

 \begin{enumerate}
    \item[-] \textbf{AdaGrasp \cite{xu2021adagrasp}:} We use a $64 \times 32 \times 64$ volumetric truncated sign distance field (TSDF), computed based on the gripper's mesh. It is first encoded with a 3D CNN and then projected with a 2D CNN to a vector embedding. We concatenate the embeddings when the gripper is fully open, half open, and fully closed.
    \item[-] \textbf{UniGrasp \cite{shao2020unigrasp}:} We first train a Pointnet-based VAE with grippers' pointcloud under random configurations. We use the 64-dim latent embedding of fully open and fully closed grippers. The input is projected to the 512-dim gripper embedding with MLP. For experiments with 10 real training grippers, the pointnet-based VAE is only trained with these 10 grippers. For experiments with procedural grippers, the pointnet-based VAE is trained with 100 randomly generated procedural grippers.
    \item[-] \textbf{PointNet++ \cite{qi2017pointnet++}:} Inspired by \cite{freiberg2024diffusionmultiembodimentgrasping}, we use PointNet++ to encode the pointcloud sampled from the gripper's mesh surface. The embedding is the concatenation of encodings when the gripper is fully open and closed.
 \end{enumerate}

 Due to the high computational cost of training \model{} with the full set of objects, we adopt a smaller scale training experiment with 453 test
 objects. Namely, we train and test on the same set of objects, while we still evaluate on the same 10 novel test grippers with the model trained on 32 procedural grippers.


\begin{table}[ht]
  \centering
  \footnotesize
  \vskip -0.08in
  \caption{Comparison with baseline gripper encoding methods used in previous work (Sec \ref{sec:gripper_encoder}). The table shows the mAUC of 453 test objects and 10 real test grippers.}
  \begin{tabular}{cccc}
    \toprule
    AdaGrasp~\cite{xu2021adagrasp} & UniGrasp~\cite{shao2020unigrasp} & PointNet++~\cite{freiberg2024diffusionmultiembodimentgrasping} & GraspGen-X \\
    \midrule
    0.432 & 0.418 & 0.349 & \textbf{0.528} \\
    \bottomrule
  \end{tabular}
  \label{tab:gripper_encoder}
  \vskip -0.1in
\end{table}

\Tableref{tab:gripper_encoder} shows the results of using other gripper encoders and inputs. Clearly, our Swept Volume heuristic shows a stronger zero-shot generalization performance on novel grippers, improving over the second best AdaGrasp (TSDF) by 25\%. This suggests that our heuristic is an efficient representation of the grasping problem. Please refer to the appendix for more results on the gripper encoder comparison.
\subsection{Ablation Study}
\label{sec:ablation_study}

~~ In this section, we present ablations on our Swept Volume parameterization and procedural gripper dataset. We follow the same experiment setup as in Sec \ref{sec:gripper_encoder}.

\begin{figure}[t]
  \centering
  \includegraphics[width=0.9\linewidth]{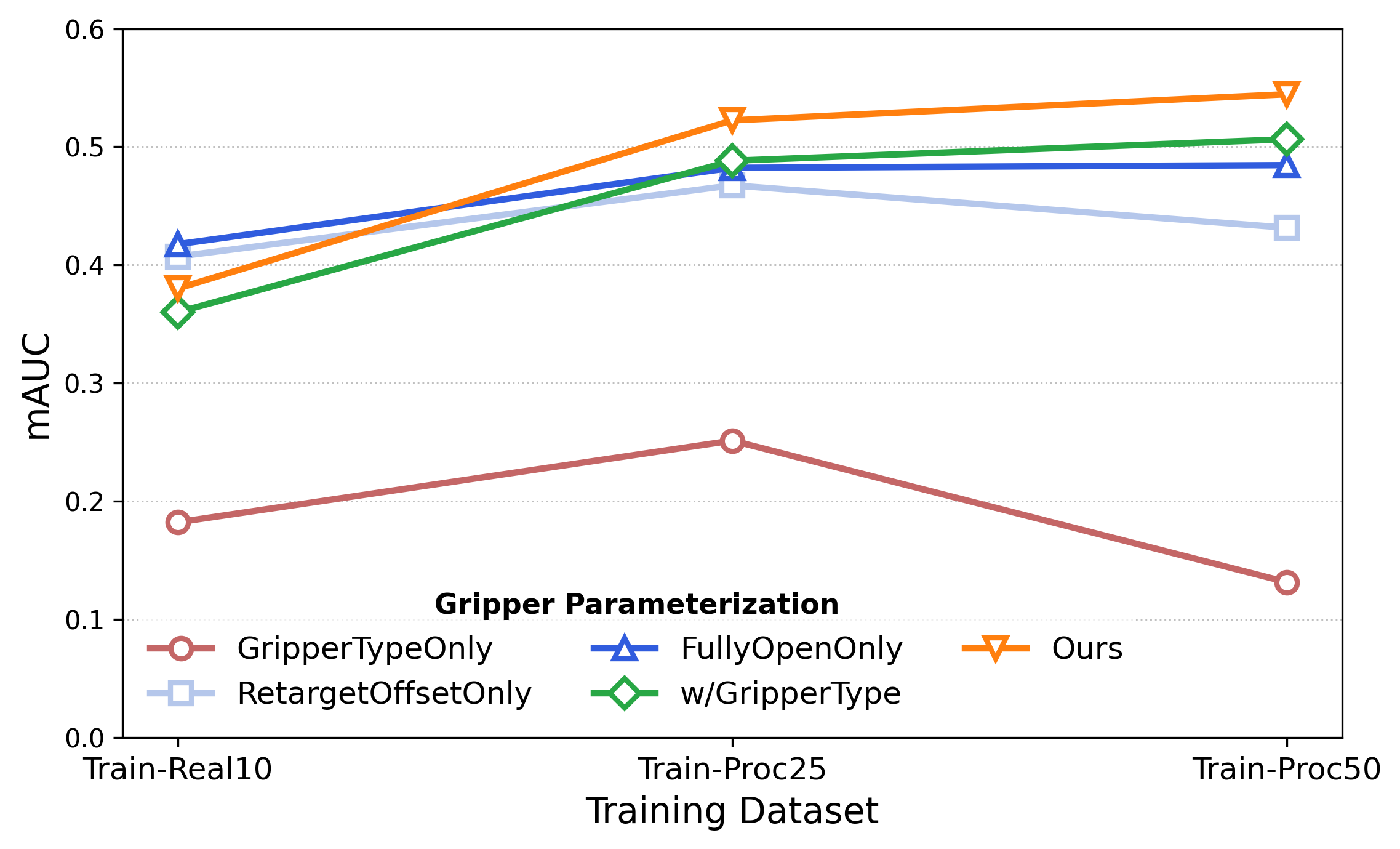}
  \vskip -0.1in
  \caption{Ablation study on \model{} gripper parameterization and gripper training sets (Sec \ref{sec:ablation_study}), evaluated on 10 test grippers.}
  \label{fig:ablation_study}
  \vskip -0.2in
\end{figure}

\textbf{Procedural Gripper Training:} To compare the performance with different training gripper distributions, we sample grasp poses and train the same model on test objects with the following set of grippers. For a fair comparison, we sample a total of 50K grasps for each object in all datasets.

 \begin{enumerate}
    \item[-] \textit{10 Real Grippers (Train-Real10).} With the 10 exclusive real grippers as training grippers, we sample 5K grasps for each gripper on each object.
    \item[-] \textit{25 procedural Grippers (Train-Proc25).} With the same 25 procedural grippers in Sec \ref{sec:method}, we sample 2K grasps for each gripper on each object.
    \item[-] \textit{50 Procedural Grippers (Train-Proc50).} We randomly generate another 50 procedural grippers with the same generator, including 20 parallel grippers, 20 2-finger revolute grippers and 10 3-finger high-dof grippers. We then sample 1K grasps for each gripper on each object.
 \end{enumerate}

Compared with training with existing real grippers, procedural gripper training yields a significant improvement for all types of gripper encodings (\Figref{fig:ablation_study}). We hypothesize that this comes from the fact that our procedural grippers provide a better coverage over test grippers, while training grippers and test grippers are largely non-overlapped and are sparsely distributed in the gripper space. Moreover, we observe that the use of more procedural grippers helps to improve performance, when using Swept Volume encoding, as well as TSDF and PointNet++ (\Figref{fig:gripper_encoder_apx}, Appendix). Here, we only use 25 procedural grippers in \model{} due to limited computation resources.
 
\textbf{Swept Volume Parameterization:} We compare on our Swept Volume parameterization with other related heuristics. All inputs are encoded with a 3-layer MLP to the gripper embedding.

 \begin{enumerate}
    \item[-] \textit{\model-GripperTypeOnly:} We condition on the gripper type, i.e., parallel, 2-finger, and 3-finger high-dof, which is a 3-dim one-hot vector.
    \item[-] \textit{\model-RetargetOffsetOnly:} We condition on the retargeting offset value, i.e., the gripper's z-axis distance between the base frame and the fingertip, a real value.
    \item[-] \textit{\model-FullyOpenOnly:} We only condition the Swept Volume when the gripper is fully open, which is a 6-dim vector.
    \item[-] \textit{\model-w/GripperType:} We use the concatenated gripper type one-hot 3-dim vector and the 12-dim Swept Volume vector (both fully open and half open) as input.
\end{enumerate}

\Figref{fig:ablation_study} shows the results of all parameterizations in our ablation study. Ours Swept Volume (12-dim vector) achieves the best performance when training with procedural grippers. GripperTypeOnly and RetargeOffsetOnly fail to learn a comparable model, due to the oversimplified gripper information in these parameterizations. We also find that using FullyOpenOnly Swept Volume performs worse than with both fully open state and half open state. This is mainly because of 2-finger revolute grippers (\Figref{fig:ablation_per_category}, Appendix). The finger of 2-finger revolute grippers such as Roboitq-2F140 and XArm Hand will move forward on the z-axis during gripper closure. Consequently, it is important to encode the information of the closing process with an additional half open Swept Volume. Interestingly, we find that conditioning on additional gripper type does not help Swept Volume. We hypothesize that when training a cross-embodiment model, it is important to share the information between different types of grippers. Thus, the additional gripper type input that separates the parameterization space degrades the performance.
\section{Real Robot Experiment}
~~~\model{} generalizes to the real world despite being only trained in simulation. Our real robot setups are shown in Fig \ref{fig:coverfig}. We demonstrate the model on two real grippers unseen by the model without any finetuning: a precise industrial manipulator of an UR10 robot equipped with a robotiq-2f-140 gripper (a 2-finger revolute gripper) and a low-cost Piper robot with its standard parallel gripper. Details of our real robot experimental setup is discussed in the Appendix.


\subsection{Evaluation on an Industrial Manipulator}
\label{sec:indust_real}
~~~\textbf{Baseline Comparisons:} We evaluate in the context of zero-shot cross-embodiment grasping, where neither \model\ nor baselines have been trained on this particular robot. We compare to two baselines: the GraspGen \cite{murali2025graspgen} model from Section \ref{sec:zero-shot} as well as AnyGrasp~\cite{fang2023anygrasp}, a recent grasping in clutter model trained on real colored point clouds of tabletop objects. For GraspGen, we retarget the predicted grasp poses meant for the Franka gripper to the Robotiq-2F140 by applying an offset along the approach direction. For the pretrained AnyGrasp model, we were unable to get consistent grasp predictions without the following post-processing steps. First, we applied a translation offset in the camera's \textit{z}-axis to match the original training dataset, which was collected at a randomized elevation/azimuth but a fixed camera depth. Second, we found better performance without applying non-maximum suppression, most likely since our motion planner \cite{curobo_report23} is proficient with batch targets. We evaluate on 12 isolated objects under 5 different poses without any clutter and on 5 objects under 3 different poses arranged on a cluttered shelf (an example is shown in the bottom left of \Figref{fig:coverfig}). 

\textbf{Discussion:} As shown in Table~\ref{tb:robot}, \model\ achieved an overall success rate of \textbf{\SI{79.0}{\percent}}, outperforming both GraspGen and AnyGrasp. \model\ performed well across both environments, though it struggled in the more challenging cluttered shelf environment. Motion planning is more difficult in clutter as most grasps would be rejected due to to kinematic infeasibility and collisions. Apart from predicting precise grasps in these new environments, the  grasp model also needs to generate grasps with high spatial coverage. This will increase the odds of having feasible grasps after a sequence of rejection sampling by the planner. Both \model\ and GraspGen are object-centric models and  hence they naturally generalized to more complicated shelf clutter with the help of SAM2. Since AnyGrasp is a scene-centric model trained only with data for tabletop clutter, it deteriorated in the out of distribution shelf scene. 

\subsection{Evaluation on a Low-Cost Robot}
~~~Low-Cost robots have the potential for democratizing robotics \cite{robotsinhomenips2018} but they come with an additional set of challenges regarding control error, actuator noise and calibration errors. We used the same \model\ model checkpoint from the previous section on the AgileX's parallel gripper. 
Here, we used object pose estimation and complete models, demonstrating that our model also generalizes to complete point cloud input. \model\ can grasp a YCB Mustard bottle and ArUCo cube with \SI{100}{\percent} success rate averaged across 10 trials, demonstrating proficiency in cross-embodiment transfer. 

\begin{table}
\footnotesize
\centering
\begin{tabular}{l|ccc}
\toprule
\textbf{Method} & \textbf{Isolated Objects} & \textbf{Clutter} & \textbf{Overall} \\
\midrule
\model (Ours) & \textbf{85.7\%} & \textbf{71.4\%} & \textbf{79.0\%} \\
GraspGen-RTG \cite{murali2025graspgen} & 73.3\% & 57.1\% & 65.2\% \\
AnyGrasp \citep{fang2023anygrasp} & 80.0\% & 42.9\% & 61.4\% \\
\bottomrule
\end{tabular}
\caption{Grasp success rate on an industrial manipulator with Robotiq-2F140 gripper (\Secref{sec:indust_real}).}
\label{tb:robot}
\vskip -0.12in
\end{table}
\section{Conclusion}

~~~Our work studies cross-embodiment 6-DOF grasping. Namely, we require the model to predict 6-DOF grasp poses not only for novel objects but also for novel gripper morphology and closing motion. In particular, our work consider grippers of 2-finger parallel grippers, 2-finger revolute grippers, and high-dof 3-finger grippers. Towards this problem, we propose \model, a diffusion-based cross-embodiment 6-dof grasping model, extending GraspGen \cite{murali2025graspgen} for cross-embodiment generalization. In \model, we use the Swept Volume heuristic to represent the gripper in the model, i.e., a 12-dim vector approximating the space that fingers will sweep during gripping. We train \model{} with procedural grippers and on the largest synthetic dataset for 6-DOF multi-embodiment grasping. Experiments suggest that our end-to-end \model{} model achieves the best performance over baseline techniques for cross-embodiment generalization. Moreover, our model serves as a good initialization for finetuning of the target novel gripper. Our ablation studies show that our gripper representation is more efficient compared to other common representations, and our procedural grippers provide a more adequate training gripper distribution. Real robot experiments show that our model, trained with only synthetic data, generalizes well to real novel grippers in 6-DOF grasping, surpassing baselines in novel objects, novel environments, and a novel embodiment.

Future work should extend our model to handle more complex grippers, e.g., training with 5-finger dexterous hands. 

\clearpage
\setcitestyle{numbers}
\bibliographystyle{ieeenat_fullname}
\bibliography{main}

\begin{thebibliography}{84}
\providecommand{\natexlab}[1]{#1}
\providecommand{\url}[1]{\texttt{#1}}
\expandafter\ifx\csname urlstyle\endcsname\relax
  \providecommand{\doi}[1]{doi: #1}\else
  \providecommand{\doi}{doi: \begingroup \urlstyle{rm}\Url}\fi

\bibitem[isa()]{isaacsim}
Isaac sim - robotics simulation and synthetic data nvidia developer.
\newblock \url{https://www.einscan.com/}.
\newblock Accessed: 2024-03-07.

\bibitem[Attarian et~al.(2023)Attarian, Asif, Liu, Hari, Garg, Gilitschenski,
  and Tompson]{Attarian_GeometryMatchingMultiEmbodimentGrasping_CoRL2023}
Maria Attarian, Muhammad~Adil Asif, Jingzhou Liu, Ruthrash Hari, Animesh Garg,
  Igor Gilitschenski, and Jonathan Tompson.
\newblock Geometry matching for multi-embodiment grasping.
\newblock In \emph{Proceedings of the 7th Conference on Robot Learning (CoRL
  2023)}, 2023.

\bibitem[Barad et~al.(2024)Barad, Orsula, Richard, Dentler, Olivares-Mendez,
  and Martinez]{barad2024graspldm}
Kuldeep Barad, Andrej Orsula, ANntoine Richard, Jan Dentler, Miguel
  Olivares-Mendez, and Carol Martinez.
\newblock Graspldm: Generative 6-dof grasp synthesis using latent diffusion
  models.
\newblock In \emph{IEEE Access}, 2024.

\bibitem[Black et~al.(2024)Black, Brown, Driess, Esmail, Equi, Finn, Fusai,
  Groom, Hausman, Ichter, Jakubczak, Jones, Ke, Levine, Li-Bell, Mothukuri,
  Nair, Pertsch, Shi, Tanner, Vuong, Walling, Wang, and
  Zhilinsky]{Black_VisionLanguageActionFlowModel2024}
Kevin Black, Noah Brown, Danny Driess, Adnan Esmail, Michael Equi, Chelsea
  Finn, Niccolò Fusai, Lachy Groom, Karol Hausman, Brian Ichter, Szymon
  Jakubczak, Tim Jones, Liyiming Ke, Sergey Levine, Adrian Li-Bell, Mohith
  Mothukuri, Suraj Nair, Karl Pertsch, Lucy~Xiaoyang Shi, James Tanner, Quan
  Vuong, Anna Walling, Haohuan Wang, and Ury Zhilinsky.
\newblock $\pi_0$: A vision-language-action flow model for general robot
  control.
\newblock arXiv preprint arXiv:2410.24164v1, 2024.

\bibitem[Bousmalis et~al.(2023)Bousmalis, Vezzani, Rao, Devin, Lee, Bauza,
  Davchev, Zhou, Gupta, Raju, Laurens, Fantacci, Dalibard, Zambelli, Martins,
  Pevceviciute, Blokzijl, Denil, Batchelor, Lampe, Parisotto, Żołna, Reed,
  Gómez~Colmenarejo, Scholz, Abdolmaleki, Groth, Regli, Sushkov, Rothörl,
  Chen, Aytar, Barker, Ortiz, Riedmiller, Springenberg, Hadsell, Nori, and
  Heess]{Bousmalis_RoboCat_TMLR2023}
Konstantinos Bousmalis, Giulia Vezzani, Dushyant Rao, Coline Devin, Alex~X.
  Lee, Maria Bauza, Todor Davchev, Yuxiang Zhou, Agrim Gupta, Akhil Raju,
  Antoine Laurens, Claudio Fantacci, Valentin Dalibard, Martina Zambelli,
  Murilo Martins, Rugile Pevceviciute, Michiel Blokzijl, Misha Denil, Nathan
  Batchelor, Thomas Lampe, Emilio Parisotto, Konrad Żołna, Scott~E. Reed,
  Sergio Gómez~Colmenarejo, Jon Scholz, Abbas Abdolmaleki, Oliver Groth,
  Jean-Baptiste Regli, Oleg Sushkov, Thomas Rothörl, José~Enrique Chen, Yusuf
  Aytar, Dave Barker, Joy Ortiz, Martin~A. Riedmiller, Jost~Tobias
  Springenberg, Raia Hadsell, Francesco Nori, and Nicolas Heess.
\newblock Robocat: A self-improving generalist agent for robotic manipulation.
\newblock \emph{Transactions on Machine Learning Research}, 12, 2023.

\bibitem[Carvalho et~al.(2024)Carvalho, Le, Jahr, Sun, Urain, Koert, and
  Peters]{carvalho2024graspdiffusionnetworklearning}
Joao Carvalho, An~T. Le, Philipp Jahr, Qiao Sun, Julen Urain, Dorothea Koert,
  and Jan Peters.
\newblock Grasp diffusion network: Learning grasp generators from partial point
  clouds with diffusion models in so(3)xr3, 2024.

\bibitem[Casas et~al.(2024)Casas, Khargonkar, Prabhakaran, and
  Xiang]{Casas_MultiGripperGrasp_IROS2024}
Luis~Felipe Casas, Ninad Khargonkar, Balakrishnan Prabhakaran, and Yu Xiang.
\newblock Multigrippergrasp: A dataset for robotic grasping from parallel jaw
  grippers to dexterous hands.
\newblock In \emph{Proceedings of the IEEE/RSJ International Conference on
  Intelligent Robots and Systems (IROS 2024)}, pages 2978--2984, 2024.

\bibitem[Chao et~al.(2021)Chao, Yang, Xiang, Molchanov, Handa, Tremblay,
  Narang, Van~Wyk, Iqbal, Birchfield, et~al.]{chao2021dexycb}
Yu-Wei Chao, Wei Yang, Yu Xiang, Pavlo Molchanov, Ankur Handa, Jonathan
  Tremblay, Yashraj~S Narang, Karl Van~Wyk, Umar Iqbal, Stan Birchfield, et~al.
\newblock Dexycb: A benchmark for capturing hand grasping of objects.
\newblock In \emph{CVPR}, 2021.

\bibitem[Chen et~al.(2025)Chen, Uy, Song, Ladhak, Murali, Qu, Birchfield,
  Blukis, and Tremblay]{chen2025spacetoolstoolaugmentedspatialreasoning}
Siyi Chen, Mikaela~Angelina Uy, Chan~Hee Song, Faisal Ladhak, Adithyavairavan
  Murali, Qing Qu, Stan Birchfield, Valts Blukis, and Jonathan Tremblay.
\newblock Spacetools: Tool-augmented spatial reasoning via double interactive
  rl, 2025.

\bibitem[Chen et~al.(2018)Chen, Murali, and
  Gupta]{Chen_Murali_Gupta_HardwareConditionedPolicies_CVPR2018}
Tao Chen, Adithyavairavan Murali, and Abhinav Gupta.
\newblock Hardware conditioned policies for multi-robot transfer learning.
\newblock In \emph{Advances in Neural Information Processing Systems 31
  (NeurIPS 2018)}, pages 9355--9366, 2018.

\bibitem[Dalal et~al.(2025)Dalal, Liu, Chen, Pathak, Zhang, and
  Salakhutdinov]{dalal2024manipgen}
Murtaza Dalal, Min Liu, Chen Chen, Deepak Pathak, Jian Zhang, and Ruslan
  Salakhutdinov.
\newblock Local policies enable zero-shot long-horizon manipulation.
\newblock In \emph{ICRA}, 2025.

\bibitem[Deshpande et~al.(2025)Deshpande, Deng, Ray, Salvador, Han, Duan, Zeng,
  Zhu, Krishna, and Hendrix]{deshpande2025graspmolmo}
Abhay Deshpande, Yuquan Deng, Arijit Ray, Jordi Salvador, Winson Han, Jiafei
  Duan, Kuo-Hao Zeng, Yuke Zhu, Ranjay Krishna, and Rose Hendrix.
\newblock Graspmolmo: Generalizable task-oriented grasping via large-scale
  synthetic data generation.
\newblock In \emph{CoRL}, 2025.

\bibitem[Devin et~al.(2017)Devin, Gupta, Darrell, Abbeel, and
  Levine]{Devin_LearningModularNeuralNetworkPolicies_ICRA2017}
Coline Devin, Abhinav Gupta, Trevor Darrell, Pieter Abbeel, and Sergey Levine.
\newblock Learning modular neural network policies for multi-task and
  multi-robot transfer.
\newblock In \emph{Proceedings of the IEEE International Conference on Robotics
  and Automation (ICRA 2017)}, pages 2169--2176, 2017.

\bibitem[Doshi et~al.(2024)Doshi, Walke, Mees, Dasari, and
  Levine]{Doshi_ScalingCrossEmbodiedLearning_CoRL2024}
Ria Doshi, Homer Walke, Oier Mees, Sudeep Dasari, and Sergey Levine.
\newblock Scaling cross-embodied learning: One policy for manipulation,
  navigation, locomotion and aviation.
\newblock In \emph{Proceedings of the Conference on Robot Learning (CoRL
  2024)}, 2024.

\bibitem[Eppner et~al.(2020)Eppner, Mousavian, and Fox]{acronym2020}
Clemens Eppner, Arsalan Mousavian, and Dieter Fox.
\newblock {ACRONYM}: A large-scale grasp dataset based on simulation.
\newblock In \emph{Under Review at ICRA 2021}, 2020.

\bibitem[Fang et~al.(2023)Fang, Wang, Fang, Gou, Liu, Yan, Liu, Xie, and
  Lu]{fang2023anygrasp}
Hao-Shu Fang, Chenxi Wang, Hongjie Fang, Minghao Gou, Jirong Liu, Hengxu Yan,
  Wenhai Liu, Yichen Xie, and Cewu Lu.
\newblock Anygrasp: Robust and efficient grasp perception in spatial and
  temporal domains.
\newblock \emph{Transactions on Robotics}, 2023.

\bibitem[Fei et~al.(2025)Fei, Wang, Luo, Gao, Cai, and Shao]{Fei_TROGrasp_2025}
Xin Fei, Zhenyu Wang, Jiayu Luo, Chongkai Gao, Zhehao Cai, and Lin Shao.
\newblock T(r,o) grasp: Efficient graph diffusion of robot-object spatial
  transformation for cross-embodiment dexterous grasping.
\newblock arXiv preprint arXiv:2510.12724, 2025.

\bibitem[Freiberg et~al.(2024)Freiberg, Qualmann, Vien, and
  Neumann]{freiberg2024diffusionmultiembodimentgrasping}
Roman Freiberg, Alexander Qualmann, Ngo~Anh Vien, and Gerhard Neumann.
\newblock Diffusion for multi-embodiment grasping.
\newblock In \emph{Robotics and Automation Letters}, 2024.

\bibitem[Freiberg et~al.(2025)Freiberg, Qualmann, Vien, and
  Neumann]{Freiberg_DiffusionForMultiEmbodimentGrasping_2025}
Roman Freiberg, Alexander Qualmann, Ngo~Anh Vien, and Gerhard Neumann.
\newblock Diffusion for multi-embodiment grasping.
\newblock \emph{IEEE Robotics and Automation Letters}, PP\penalty0
  (99):\penalty0 1--8, 2025.

\bibitem[Fu et~al.(2025)Fu, Yu, El-Refai, Kou, Xue, Huang, Xiao, Wang, Li, Shi,
  Wu, Sastry, Zhu, Goldberg, and Fan]{fu2025capx}
Max Fu, Justin Yu, Karim El-Refai, Ethan Kou, Haoru Xue, Huang Huang, Wenli
  Xiao, Guanzhi Wang, Fei-Fei Li, Guanya Shi, Jiajun Wu, Shankar Sastry, Yuke
  Zhu, Ken Goldberg, and Jim Fan.
\newblock {CaP-X}: A framework for benchmarking and improving coding agents for
  robot manipulation.
\newblock \emph{arXiv preprint arXiv:2603.22435}, 2025.

\bibitem[Gupta et~al.(2018{\natexlab{a}})Gupta, Murali, Gandhi, and
  Pinto]{Gupta_Murali_Gandhi_Pinto_RobotLearningInHomes_NeurIPS2018}
Abhinav Gupta, Adithyavairavan Murali, Dhiraj Gandhi, and Lerrel Pinto.
\newblock Robot learning in homes: Improving generalization and reducing
  dataset bias.
\newblock In \emph{Advances in Neural Information Processing Systems 31
  (NeurIPS 2018)}, 2018{\natexlab{a}}.

\bibitem[Gupta et~al.(2018{\natexlab{b}})Gupta, Murali, Gandhi, and
  Pinto]{robotsinhomenips2018}
Abhinav Gupta, Adithyavairavan Murali, Dhiraj Gandhi, and Lerrel Pinto.
\newblock Robot learning in homes: Improving generalization and reducing
  dataset bias.
\newblock \emph{NeurIPS}, 2018{\natexlab{b}}.

\bibitem[Hampali et~al.(2020)Hampali, Rad, Oberweger, and
  Lepetit]{hampali2020honnotate-ho3d}
Shreyas Hampali, Mahdi Rad, Markus Oberweger, and Vincent Lepetit.
\newblock Honnotate: A method for 3d annotation of hand and object poses.
\newblock In \emph{CVPR}, 2020.

\bibitem[Handa et~al.(2020)Handa, Van~Wyk, Yang, Liang, Chao, Wan, Birchfield,
  Ratliff, and Fox]{Handa_DexPilot_ICRA2020}
Ankur Handa, Karl Van~Wyk, Wei Yang, Jacky Liang, Yu-Wei Chao, Qian Wan, Stan
  Birchfield, Nathan Ratliff, and Dieter Fox.
\newblock Dexpilot: Vision-based teleoperation of dexterous robotic hand-arm
  system.
\newblock In \emph{Proceedings of the IEEE International Conference on Robotics
  and Automation (ICRA 2020)}, pages 9164--9170, 2020.

\bibitem[Hu et~al.(2022)Hu, Huang, Rybkin, and
  Jayaraman]{Hu_KnowThyself_ICLR2022}
Edward~S. Hu, Kun Huang, Oleh Rybkin, and Dinesh Jayaraman.
\newblock Know thyself: Transferable visual control policies through
  robot-awareness.
\newblock In \emph{Proceedings of the International Conference on Learning
  Representations (ICLR 2022)}, 2022.

\bibitem[Huang et~al.(2020)Huang, Mordatch, and
  Pathak]{Huang_OnePolicyToControlThemAll_ICML2020}
Wenlong Huang, Igor Mordatch, and Deepak Pathak.
\newblock One policy to control them all: Shared modular policies for
  agent-agnostic control.
\newblock In \emph{Proceedings of the International Conference on Machine
  Learning (ICML 2020)}, pages 4455--4464, 2020.

\bibitem[Huang et~al.(2024)Huang, Wang, Li, Zhang, and Fei-Fei]{huang2024rekep}
Wenlong Huang, Chen Wang, Yunzhu Li, Ruohan Zhang, and Li Fei-Fei.
\newblock Rekep: Spatio-temporal reasoning of relational keypoint constraints
  for robotic manipulation.
\newblock In \emph{CoRL}, 2024.

\bibitem[Joshi et~al.(2025)Joshi, Han, Nugent, Saez-Diez, Zuo, Liu, Wen,
  Alexandropoulos, Kayan, Calveri, Sun, Liu, Shao, Raistrick, and
  Deng]{joshi2025infinigensim}
Abhishek Joshi, Beining Han, Jack Nugent, Max~Gonzalez Saez-Diez, Yiming Zuo,
  Jonathan Liu, Hongyu Wen, Stamatis Alexandropoulos, Karhan Kayan, Anna
  Calveri, Tao Sun, Gaowen Liu, Yi Shao, Alexander Raistrick, and Jia Deng.
\newblock Procedural generation of articulated simulation-ready assets, 2025.

\bibitem[Levine et~al.(2016)Levine, Pastor, Krizhevsky, and
  Quillen]{Levine2016LearningHandEye}
Sergey Levine, Peter Pastor, Alex Krizhevsky, and Deirdre Quillen.
\newblock Learning hand-eye coordination for robotic grasping with deep
  learning and large-scale data collection, 2016.

\bibitem[Li et~al.(2025{\natexlab{a}})Li, Sun, Hu, Ta, Barry, Konidaris, and
  Fu]{li2025novaflow}
Hongyu Li, Lingfeng Sun, Yafei Hu, Duy Ta, Jennifer Barry, George Konidaris,
  and Jiahui Fu.
\newblock Novaflow: Zero-shot manipulation via actionable flow from generated
  videos.
\newblock \emph{arXiv preprint arXiv:2510.08568}, 2025{\natexlab{a}}.

\bibitem[Li et~al.(2025{\natexlab{b}})Li, Zhu, Tang, Wen, Zhu, Liu, Li, Cheng,
  Peng, Peng, and Feng]{Li2025CoAVLA}
Jinming Li, Yichen Zhu, Zhibin Tang, Junjie Wen, Minjie Zhu, Xiaoyu Liu,
  Chengmeng Li, Ran Cheng, Yaxin Peng, Yan Peng, and Feifei Feng.
\newblock Coa-vla: Improving vision-language-action models via visual-textual
  chain-of-affordance.
\newblock In \emph{Proceedings of the IEEE/CVF International Conference on
  Computer Vision (ICCV 2025)}, 2025{\natexlab{b}}.
\newblock arXiv preprint arXiv:2412.20451.

\bibitem[Li et~al.(2023)Li, Liu, Li, Geng, Zhu, Yang, and
  Huang]{Li_GenDexGrasp_ICRA2023}
Puhao Li, Tengyu Liu, Yuyang Li, Yiran Geng, Yixin Zhu, Yaodong Yang, and
  Siyuan Huang.
\newblock Gendexgrasp: Generalizable dexterous grasping.
\newblock In \emph{Proceedings of the IEEE International Conference on Robotics
  and Automation (ICRA 2023)}, 2023.

\bibitem[Liang et~al.(2019)Liang, Ma, Li, Gorner, Tang, Fang, Sun, and
  Zhang]{PointNetGPD2019}
Hongzhuo Liang, Xiaojian Ma, Shuang Li, Michael Gorner, Song Tang, Bin Fang,
  Fuchun Sun, and Jianwei Zhang.
\newblock Pointnetgpd: Detecting grasp configurations from point sets.
\newblock In \emph{ICRA}, 2019.

\bibitem[Lim et~al.(2024)Lim, Kim, Kim, Lee, and Park]{lim2024equigraspflow}
Byeongdo Lim, Jongmin Kim, Jihwan Kim, Yonghyeon Lee, and Frank~C Park.
\newblock Equigraspflow: Se(3)-equivariant 6-dof grasp pose generative flows.
\newblock In \emph{CoRL}, 2024.

\bibitem[Liu et~al.(2020)Liu, Pan, Xu, Ganguly, and Manocha]{liu2020deep-ddg}
Min Liu, Zherong Pan, Kai Xu, Kanishka Ganguly, and Dinesh Manocha.
\newblock Deep differentiable grasp planner for high-dof grippers.
\newblock \emph{arXiv preprint arXiv:2002.01530}, 2020.

\bibitem[Liu et~al.(2024)Liu, Orru, Paxton, Shafiullah, and
  Pinto]{liu2024okrobot}
Peiqi Liu, Yaswanth Orru, Chris Paxton, Nur Muhammad~Mahi Shafiullah, and
  Lerrel Pinto.
\newblock Ok-robot: What really matters in integrating open-knowledge models
  for robotics.
\newblock \emph{preprint arXiv:2401.12202}, 2024.

\bibitem[Lum et~al.(2024)Lum, Li, Culbertson, Srinivasan, Ames, Schwager, and
  Bohg]{lum2024get}
Tyler Ga~Wei Lum, Albert~H. Li, Preston Culbertson, Krishnan Srinivasan, Aaron
  Ames, Mac Schwager, and Jeannette Bohg.
\newblock Get a grip: Multi-finger grasp evaluation at scale enables robust
  sim-to-real transfer.
\newblock In \emph{CoRL}, 2024.

\bibitem[Macklin et~al.(2014)Macklin, M{\"u}ller, Chentanez, and
  Kim]{macklin2014unified-flex-simulator}
Miles Macklin, Matthias M{\"u}ller, Nuttapong Chentanez, and Tae-Yong Kim.
\newblock Unified particle physics for real-time applications.
\newblock \emph{ACM Transactions on Graphics (TOG)}, 33\penalty0 (4):\penalty0
  1--12, 2014.

\bibitem[Mahler et~al.(2017)Mahler, Liang, Niyaz, Laskey, Doan, Liu, Ojea, and
  Goldberg]{mahler2017dexnet2}
Jeffrey Mahler, Jacky Liang, Sherdil Niyaz, Michael Laskey, Richard Doan, Xinyu
  Liu, Juan~Aparicio Ojea, and Ken Goldberg.
\newblock Dex-net 2.0: Deep learning to plan robust grasps with synthetic point
  clouds and analytic grasp metrics.
\newblock \emph{arXiv preprint arXiv:1703.09312}, 2017.

\bibitem[Millane et~al.(2024{\natexlab{a}})Millane, Oleynikova, Wirbel,
  Steiner, Ramasamy, Tingdahl, and Siegwart]{Millane_nvblox_ICRA2024}
Alexander Millane, Helen Oleynikova, Emilie Wirbel, Remo Steiner, Vikram
  Ramasamy, David Tingdahl, and Roland Siegwart.
\newblock nvblox: Gpu-accelerated incremental signed distance field mapping.
\newblock In \emph{Proceedings of the IEEE International Conference on Robotics
  and Automation}, pages 2698--2705, 2024{\natexlab{a}}.

\bibitem[Millane et~al.(2024{\natexlab{b}})Millane, Oleynikova, Wirbel,
  Steiner, Ramasamy, Tingdahl, and Siegwart]{millane2024nvblox}
Alexander Millane, Helen Oleynikova, Emilie Wirbel, Remo Steiner, Vikram
  Ramasamy, David Tingdahl, and Roland Siegwart.
\newblock nvblox: Gpu-accelerated incremental signed distance field mapping,
  2024{\natexlab{b}}.

\bibitem[Morrison et~al.(2020)Morrison, Corke, and Leitner]{morrison2020egad}
Douglas Morrison, Peter Corke, and J{\"u}rgen Leitner.
\newblock Egad! an evolved grasping analysis dataset for diversity and
  reproducibility in robotic manipulation.
\newblock \emph{IEEE Robotics and Automation Letters}, 5\penalty0 (3):\penalty0
  4368--4375, 2020.

\bibitem[Mousavian et~al.(2019)Mousavian, Eppner, and
  Fox]{mousavian2019-6dofgraspnet}
Arsalan Mousavian, Clemens Eppner, and Dieter Fox.
\newblock 6-dof graspnet: Variational grasp generation for object manipulation.
\newblock In \emph{ICCV}, 2019.

\bibitem[Murali et~al.(2020{\natexlab{a}})Murali, Liu, Marino, Chernova, and
  Gupta]{murali2020taskgrasp}
Adithyavairavan Murali, Weiyu Liu, Kenneth Marino, Sonia Chernova, and Abhinav
  Gupta.
\newblock Same object, different grasps: Data and semantic knowledge for
  task-oriented grasping.
\newblock In \emph{CoRL}, 2020{\natexlab{a}}.

\bibitem[Murali et~al.(2020{\natexlab{b}})Murali, Mousavian, Eppner, Paxton,
  and Fox]{Murali2020CollisionNet}
Adithyavairavan Murali, Arsalan Mousavian, Clemens Eppner, Chris Paxton, and
  Dieter Fox.
\newblock 6-dof grasping for target-driven object manipulation in clutter.
\newblock In \emph{ICRA}, 2020{\natexlab{b}}.

\bibitem[Murali et~al.(2025)Murali, Sundaralingam, Chao, Yamada, Yuan, Carlson,
  Ramos, Birchfield, Fox, and Eppner]{murali2025graspgen}
Adithyavairavan Murali, Balakumar Sundaralingam, Yu-Wei Chao, Jun Yamada,
  Wentao Yuan, Mark Carlson, Fabio Ramos, Stan Birchfield, Dieter Fox, and
  Clemens Eppner.
\newblock Graspgen: A diffusion-based framework for 6-dof grasping with
  on-generator training.
\newblock \emph{arXiv preprint arXiv:2507.13097}, 2025.

\bibitem[Newbury et~al.(2023)Newbury, Gu, Chumbley, Mousavian, Eppner, Leitner,
  Bohg, Morales, Asfour, Kragic, et~al.]{newbury2023deepgraspsurvey}
Rhys Newbury, Morris Gu, Lachlan Chumbley, Arsalan Mousavian, Clemens Eppner,
  J{\"u}rgen Leitner, Jeannette Bohg, Antonio Morales, Tamim Asfour, Danica
  Kragic, et~al.
\newblock Deep learning approaches to grasp synthesis: A review.
\newblock \emph{IEEE Transactions on Robotics}, 2023.

\bibitem[NVIDIA(2023)]{nvidia2023-isaac-sim}
NVIDIA.
\newblock Nvidia isaac sim, 2023.

\bibitem[Qi et~al.(2017)Qi, Yi, Su, and Guibas]{qi2017pointnet++}
Charles~Ruizhongtai Qi, Li Yi, Hao Su, and Leonidas~J Guibas.
\newblock Pointnet++: Deep hierarchical feature learning on point sets in a
  metric space.
\newblock \emph{Advances in neural information processing systems}, 30, 2017.

\bibitem[Qin et~al.(2023)Qin, Yang, Huang, Van~Wyk, Su, Wang, Chao, and
  Fox]{Qin_AnyTeleop_2023}
Yuzhe Qin, Wei Yang, Binghao Huang, Karl Van~Wyk, Hao Su, Xiaolong Wang, Yu-Wei
  Chao, and Dieter Fox.
\newblock Anyteleop: A general vision-based dexterous robot arm-hand
  teleoperation system.
\newblock arXiv preprint arXiv:2307.04577v1, 2023.

\bibitem[Raistrick et~al.(2023)Raistrick, Lipson, Ma, Mei, Wang, Zuo, Kayan,
  Wen, Han, Wang, et~al.]{raistrick2023infinigen}
Alexander Raistrick, Lahav Lipson, Zeyu Ma, Lingjie Mei, Mingzhe Wang, Yiming
  Zuo, Karhan Kayan, Hongyu Wen, Beining Han, Yihan Wang, et~al.
\newblock Infinite photorealistic worlds using procedural generation.
\newblock In \emph{Proceedings of the IEEE/CVF Conference on Computer Vision
  and Pattern Recognition}, pages 12630--12641, 2023.

\bibitem[Raistrick et~al.(2024)Raistrick, Lingjie, Karhan, David, Yiming,
  Beining, Hongyu, Meenal, Alexandropoulos, Lahav, Zeyu, and
  Jia]{raistrick2024infinigen_indoors}
Alexander Raistrick, Mei Lingjie, Kaan~Kayan Karhan, Yan David, Zuo Yiming, Han
  Beining, Wen Hongyu, Parakh Meenal, Stamatis Alexandropoulos, Lipson Lahav,
  Ma Zeyu, and Deng Jia.
\newblock Infinite indoors: Photorealistic indoor scenes using procedural
  generation.
\newblock In \emph{Proceedings of the IEEE/CVF Conference on Computer Vision
  and Pattern Recognition}, 2024.

\bibitem[Ravi et~al.(2024)Ravi, Gabeur, Hu, Hu, Ryali, Ma, Khedr, R{\"a}dle,
  Rolland, Gustafson, Mintun, Pan, Alwala, Carion, Wu, Girshick, Doll{\'a}r,
  and Feichtenhofer]{ravi2024sam2}
Nikhila Ravi, Valentin Gabeur, Yuan-Ting Hu, Ronghang Hu, Chaitanya Ryali,
  Tengyu Ma, Haitham Khedr, Roman R{\"a}dle, Chloe Rolland, Laura Gustafson,
  Eric Mintun, Junting Pan, Kalyan~Vasudev Alwala, Nicolas Carion, Chao-Yuan
  Wu, Ross Girshick, Piotr Doll{\'a}r, and Christoph Feichtenhofer.
\newblock Sam 2: Segment anything in images and videos.
\newblock \emph{arXiv preprint arXiv:2408.00714}, 2024.

\bibitem[Salhotra et~al.(2023)Salhotra, Liu, and
  Sukhatme]{Salhotra_LearningRobotManipulation_CoRL2023}
Gaurav Salhotra, I-Chun~Arthur Liu, and Gaurav~S. Sukhatme.
\newblock Learning robot manipulation from cross-morphology demonstration.
\newblock In \emph{Proceedings of the 7th Annual Conference on Robot Learning
  (CoRL 2023)}, 2023.

\bibitem[Shao et~al.(2020)Shao, Ferreira, Jorda, Nambiar, Luo, Solowjow, Ojea,
  Khatib, and Bohg]{shao2020unigrasp}
Lin Shao, Fabio Ferreira, Mikael Jorda, Varun Nambiar, Jianlan Luo, Eugen
  Solowjow, Juan~Aparicio Ojea, Oussama Khatib, and Jeannette Bohg.
\newblock Unigrasp: Learning a unified model to grasp with multifingered
  robotic hands.
\newblock \emph{IEEE Robotics and Automation Letters}, 5\penalty0 (2):\penalty0
  2286--2293, 2020.

\bibitem[Shen et~al.(2026)Shen, Kumar, Chintalapudi, Wang, Watson, Hu, Cao,
  Jayaraman, Kaelbling, and Lozano-P{\'e}rez]{shen2026tiptop}
William Shen, Nishanth Kumar, Sahit Chintalapudi, Jie Wang, Christopher Watson,
  Edward~S. Hu, Jing Cao, Dinesh Jayaraman, Leslie~Pack Kaelbling, and
  Tom{\'a}s Lozano-P{\'e}rez.
\newblock {TiPToP}: A modular open-vocabulary planning system for robotic
  manipulation.
\newblock \emph{arXiv preprint arXiv:2603.09971}, 2026.

\bibitem[Shi et~al.(2025)Shi, Yang, Chao, Wan, Shao, Lei, Qian, Le, Chaudhari,
  Daniilidis, Wen, and Jayaraman]{Shi2025Maestro}
Junyao Shi, Rujia Yang, Kaitian Chao, Selina~Bingqing Wan, Yifei Shao, Jiahui
  Lei, Jianing Qian, Long Le, Pratik Chaudhari, Kostas Daniilidis, Chuan Wen,
  and Dinesh Jayaraman.
\newblock Maestro: Orchestrating robotics modules with vision-language models
  for zero-shot generalist robots.
\newblock \emph{arXiv preprint arXiv:2511.00917}, 2025.

\bibitem[Song et~al.(2024)Song, Li, and Detry]{song2024implicitgraspdiffusion}
Pinhao Song, Pengteng Li, and Renaud Detry.
\newblock Implicit grasp diffusion: Bridging the gap between dense prediction
  and sampling-based grasping.
\newblock In \emph{CoRL}, 2024.

\bibitem[Sundaralingam et~al.(2023{\natexlab{a}})Sundaralingam, Hari, Fishman,
  Garrett, Van~Wyk, Blukis, Millane, Oleynikova, Handa, Ramos,
  et~al.]{sundaralingam2023curobo}
Balakumar Sundaralingam, Siva Kumar~Sastry Hari, Adam Fishman, Caelan Garrett,
  Karl Van~Wyk, Valts Blukis, Alexander Millane, Helen Oleynikova, Ankur Handa,
  Fabio Ramos, et~al.
\newblock Curobo: Parallelized collision-free robot motion generation.
\newblock In \emph{2023 IEEE International Conference on Robotics and
  Automation (ICRA)}, pages 8112--8119. IEEE, 2023{\natexlab{a}}.

\bibitem[Sundaralingam et~al.(2023{\natexlab{b}})Sundaralingam, Hari, Fishman,
  Garrett, Wyk, Blukis, Millane, Oleynikova, Handa, Ramos, Ratliff, and
  Fox]{curobo_report23}
Balakumar Sundaralingam, Siva Kumar~Sastry Hari, Adam Fishman, Caelan Garrett,
  Karl~Van Wyk, Valts Blukis, Alexander Millane, Helen Oleynikova, Ankur Handa,
  Fabio Ramos, Nathan Ratliff, and Dieter Fox.
\newblock curobo: Parallelized collision-free minimum-jerk robot motion
  generation.
\newblock In \emph{ICRA}, 2023{\natexlab{b}}.

\bibitem[Sundermeyer et~al.(2021)Sundermeyer, Mousavian, Triebel, and
  Fox]{sundermeyer2021contact}
Martin Sundermeyer, Arsalan Mousavian, Rudolph Triebel, and Dieter Fox.
\newblock Contact-graspnet: Efficient 6-dof grasp generation in cluttered
  scenes.
\newblock In \emph{2021 IEEE International Conference on Robotics and
  Automation (ICRA)}, pages 13438--13444. IEEE, 2021.

\bibitem[Tang et~al.(2023)Tang, Huang, Ge, Liu, and Zhang]{tang2023graspgpt}
Chao Tang, Dehao Huang, Wenqi Ge, Weiyu Liu, and Hong Zhang.
\newblock Graspgpt: Leveraging semantic knowledge from a large language model
  for task-oriented grasping.
\newblock \emph{RAL}, 2023.

\bibitem[ten Pas et~al.(2017)ten Pas, Gualtieri, Saenko, and
  Platt]{tenPas2017GraspPoseDetection}
Andreas ten Pas, Marcus Gualtieri, Kate Saenko, and Robert Platt.
\newblock Grasp pose detection in point clouds.
\newblock \emph{The International Journal of Robotics Research}, 36\penalty0
  (13–14):\penalty0 1455--1473, 2017.

\bibitem[Tobin et~al.(2018)Tobin, Biewald, Duan, Andrychowicz, Handa, Kumar,
  McGrew, Ray, Schneider, Welinder, Zaremba, and Abbeel]{tobin2018grasp}
Josh Tobin, Lukas Biewald, Rocky Duan, Marcin Andrychowicz, Ankur Handa, Vikash
  Kumar, Bob McGrew, Alex Ray, Jonas Schneider, Peter Welinder, Wojciech
  Zaremba, and Pieter Abbeel.
\newblock Domain randomization and generative models for robotic grasping.
\newblock In \emph{IROS}, 2018.

\bibitem[Turpin et~al.(2023)Turpin, Zhong, Zhang, Zhu, Liu, Singh, Heiden,
  Macklin, Tsogkas, Dickinson, et~al.]{turpin2023fast-graspd}
Dylan Turpin, Tao Zhong, Shutong Zhang, Guanglei Zhu, Jingzhou Liu, Ritvik
  Singh, Eric Heiden, Miles Macklin, Stavros Tsogkas, Sven Dickinson, et~al.
\newblock Fast-grasp'd: Dexterous multi-finger grasp generation through
  differentiable simulation.
\newblock \emph{arXiv:2306.08132}, 2023.

\bibitem[Urain et~al.(2023)Urain, Funk, Peters, and
  Chalvatzaki]{urain2022se3dif}
Julen Urain, Niklas Funk, Jan Peters, and Georgia Chalvatzaki.
\newblock Se(3)-diffusionfields: Learning smooth cost functions for joint grasp
  and motion optimization through diffusion.
\newblock \emph{ICRA}, 2023.

\bibitem[Wang et~al.(2023)Wang, Zhang, Chen, Xu, Li, Liu, and
  Wang]{wang2023dexgraspnet}
Ruicheng Wang, Jialiang Zhang, Jiayi Chen, Yinzhen Xu, Puhao Li, Tengyu Liu,
  and He Wang.
\newblock Dexgraspnet: A large-scale robotic dexterous grasp dataset for
  general objects based on simulation.
\newblock In \emph{ICRA}, 2023.

\bibitem[Wang and Xu(2024)]{Wang_TransferringGraspingAcrossGrippers_TRO2024}
Xianli Wang and Qingsong Xu.
\newblock Transferring grasping across grippers: Learning–optimization hybrid
  framework for generalized planar grasp generation.
\newblock \emph{IEEE Transactions on Robotics}, 2024.

\bibitem[Wang et~al.(2020)Wang, Garrett, Kaelbling, and
  Lozano-Pérez]{Wang_LearningCompositionalModels_2020}
Zi Wang, Caelan~Reed Garrett, Leslie~Pack Kaelbling, and Tomas Lozano-Pérez.
\newblock Learning compositional models of robot skills for task and motion
  planning.
\newblock \emph{The International Journal of Robotics Research}, 2020.

\bibitem[Wei et~al.(2022)Wei, Liu, Ling, and Su]{wei2022coacd}
Xinyue Wei, Minghua Liu, Zhan Ling, and Hao Su.
\newblock Approximate convex decomposition for 3d meshes with collision-aware
  concavity and tree search.
\newblock \emph{ACM Transactions on Graphics (TOG)}, 41\penalty0 (4):\penalty0
  1--18, 2022.

\bibitem[Wei et~al.(2025)Wei, Xu, Guo, Hou, Gao, Cai, Luo, and
  Shao]{Wei_DROGrasp_ICRA2025}
Zhenyu Wei, Zhixuan Xu, Jingxiang Guo, Yiwen Hou, Chongkai Gao, Zhehao Cai,
  Jiayu Luo, and Lin Shao.
\newblock D(r,o) grasp: A unified representation of robot and object
  interaction for cross-embodiment dexterous grasping.
\newblock In \emph{Proceedings of the IEEE International Conference on Robotics
  and Automation (ICRA 2025)}, 2025.

\bibitem[Wen et~al.(2023)Wen, Yang, Kautz, and
  Birchfield]{wen2023foundationpose}
Bowen Wen, Wei Yang, Jan Kautz, and Stan Birchfield.
\newblock Foundationpose: Unified 6d pose estimation and tracking of novel
  objects.
\newblock \emph{arXiv preprint arXiv:2312.08344}, 2023.

\bibitem[Wen et~al.(2025)Wen, Trepte, Aribido, Kautz, Gallo, and
  Birchfield]{wen2025foundationstereo}
Bowen Wen, Matthew Trepte, Joseph Aribido, Jan Kautz, Orazio Gallo, and Stan
  Birchfield.
\newblock Foundationstereo: Zero-shot stereo matching.
\newblock In \emph{Proceedings of the Computer Vision and Pattern Recognition
  Conference}, pages 5249--5260, 2025.

\bibitem[Weng et~al.(2024)Weng, Lu, Kragic, and Lundell]{weng2024dexdiffuser}
Zehang Weng, Haofei Lu, Danica Kragic, and Jens Lundell.
\newblock Dexdiffuser: Generating dexterous grasps with diffusion models.
\newblock \emph{Robotics and Automation Letters}, 2024.

\bibitem[Wu et~al.(2024)Wu, Jiang, Wang, Liu, Liu, Qiao, Ouyang, He, and
  Zhao]{wu2024ptv3}
Xiaoyang Wu, Li Jiang, Peng-Shuai Wang, Zhijian Liu, Xihui Liu, Yu Qiao, Wanli
  Ouyang, Tong He, and Hengshuang Zhao.
\newblock Point transformer v3: Simpler, faster, stronger.
\newblock In \emph{CVPR}, 2024.

\bibitem[Wu et~al.(2023)Wu, Wang, and
  Wang]{wu2023learning-dex-human-affordance}
Yueh-Hua Wu, Jiashun Wang, and Xiaolong Wang.
\newblock Learning generalizable dexterous manipulation from human grasp
  affordance.
\newblock In \emph{CoRL}, 2023.

\bibitem[X-Team(2024)]{ONeill_OpenXEmbodiment_ICRA2024}
Open X-Team.
\newblock Open x-embodiment: Robotic learning datasets and rt-x models.
\newblock In \emph{Proceedings of the IEEE International Conference on Robotics
  and Automation (ICRA 2024)}, pages 6892--6903, 2024.

\bibitem[Xie et~al.(2025)Xie, Chen, Tang, Hu, Yang, and
  Wang]{xie2024rethinking6dofgraspdetection}
Pengwei Xie, Siang Chen, Wei Tang, Dingchang Hu, Wenming Yang, and Guijin Wang.
\newblock Rethinking 6-dof grasp detection: A flexible framework for
  high-quality grasping.
\newblock \emph{Pattern Recognition}, 2025.

\bibitem[Xu et~al.(2021)Xu, Qi, Agrawal, and Song]{xu2021adagrasp}
Zhenjia Xu, Beichun Qi, Shubham Agrawal, and Shuran Song.
\newblock Adagrasp: Learning an adaptive gripper-aware grasping policy.
\newblock In \emph{2021 IEEE International Conference on Robotics and
  Automation (ICRA)}, pages 4620--4626. IEEE, 2021.

\bibitem[Yamada et~al.(2025)Yamada, Murali, Mandlekar, Eppner, Posner, and
  Sundaralingam]{Yamada_GraspMPC_2025}
Jun Yamada, Adithyavairavan Murali, Ajay Mandlekar, Clemens Eppner, Ingmar
  Posner, and Balakumar Sundaralingam.
\newblock Grasp-mpc: Closed-loop visual grasping via value-guided model
  predictive control.
\newblock arXiv preprint arXiv:2509.06201v1, 2025.

\bibitem[Yu et~al.(2017)Yu, Tan, Liu, and
  Turk]{Yu_LearningUniversalPolicy_RSS2017}
Wenhao Yu, Jie Tan, C.~Karen Liu, and Greg Turk.
\newblock Preparing for the unknown: Learning a universal policy with online
  system identification.
\newblock In \emph{Proceedings of the Robotics: Science and Systems (RSS
  2017)}, 2017.

\bibitem[Yuan et~al.(2023)Yuan, Murali, and Mousavian]{yuan2023m2t2}
Wentao Yuan, Adithyavairavan Murali, and Arsalan Mousavian.
\newblock M2t2: Multi-task masked transformer for object-centric pick and
  place.
\newblock In \emph{7th Annual Conference on Robot Learning}.
  https://openreview. net/forum? id= 6zGpfOBImD, 2023.

\bibitem[Yuan et~al.(2025)Yuan, Duan, Blukis, Pumacay, Krishna, Murali,
  Mousavian, and Fox]{Yuan2025RoboPoint}
Wentao Yuan, Jiafei Duan, Valts Blukis, Wilbert Pumacay, Ranjay Krishna,
  Adithyavairavan Murali, Arsalan Mousavian, and Dieter Fox.
\newblock Robopoint: A vision-language model for spatial affordance prediction
  for robotics.
\newblock In \emph{Proceedings of the 8th Conference on Robot Learning (CoRL
  2024)}, pages 4005--4020. PMLR, 2025.

\bibitem[Zakka et~al.(2021)Zakka, Zeng, Florence, Tompson, Bohg, and
  Dwibedi]{Zakka_XIRL_CoRL2021}
Kevin Zakka, Andy Zeng, Pete Florence, Jonathan Tompson, Jeannette Bohg, and
  Debidatta Dwibedi.
\newblock Xirl: Cross-embodiment inverse reinforcement learning.
\newblock In \emph{Proceedings of the Conference on Robot Learning (CoRL
  2021)}, pages 537--546, 2021.

\end{thebibliography}

\clearpage
\maketitlesupplementary


\section{Procedural Gripper Generator}
\label{appendix-procgripper}
~~~We design procedural gripper generators with Infinigen Articulated \cite{joshi2025infinigensim}. Infinigen Articulated is the extension of Infinigen \cite{raistrick2023infinigen, raistrick2024infinigen_indoors}, which supports the design of procedural articulated objects with the Blender Geometry Node. For all Infinigen objects, each generator consists of randomized mathematical functions, without using any existing data / meshes. Here, we design a generator for each category of parallel grippers, 2-finger revolute grippers, and 3-finger high-dof grippers.

\textbf{2-Finger Parallel Gripper.} Each parallel gripper consists of 3 links, i.e., gripper base, left finger, right finger, and 2 prismatic joints, i.e., the left finger joint and the right finger joint. The gripper base consists of a cube and a cylinder with a random value to control the ratio. We also randomize the base geometry in terms of its height, width, and depth. We create fingers as cubic objects and are modified based on the ratio of fingertip width to the finger bottom width, and its tilting ratio. In half of all samples, we add additional meshes of square, round, or triangle cylinders at the fingertip.

\textbf{2-Finger Revolute Gripper.} Each 2-finger revolute gripper consists of 5 links, i.e., gripper base, left mid finger, left top finger, right mid finger, right top finger, and 4 revolute joints, i.e., left mid finger joint, left top finger joins, right mid finger joint, and right top finger joint. The gripper base is randomized in dimension and in the ratio between the base top and base bottom. The mid finger links and the top finger links are cubic objects. For mid fingers, we randomly add an outer finger like in Robotiq-2F and OnRobot-RG grippers. For the top fingers, we randomly add round / square cylinders at fingertips. There are two modes for gripper closing motion. In the first mode, the right and left top finger links are always parallel to each other. Thus, the top finger joint and the mid finger joint rotate in a ratio of $1:-1$. In the second mode, all top finger joints and the mid finger joints rotate in a ratio of $1:1$, where the gripper fingers will close like a pinch gripper.

\textbf{3-Finger High-DOF Grippers.} Each 3-finger high-dof gripper consists of the gripper base and 3 2-joint / 3-joint fingers. We create a cubic gripper base with randomized dimensions and wrist-to-palm ratio. We attach two fingers at the top of the palm and one finger at the center of the wrist. All fingers are stretching in x-axis when it is open. The total DOFs with 2-joint fingers is 6 and the total DOF with 3-joint fingers is 9. All finger links are cubic objects with random width, depth, and height. Moreover, we randomly change the orientation of the two fingers at the top of the palm around the z-axis to mimic the potential side rotation that real 3-finger hands have (e.g., Robotiq-3F). For the gripper closing motion, all joints follow a linearly interpolated trajectory from the fully open state to the fully closed state.

\begin{figure*}[t]
  \centering
  \includegraphics[width=\linewidth]{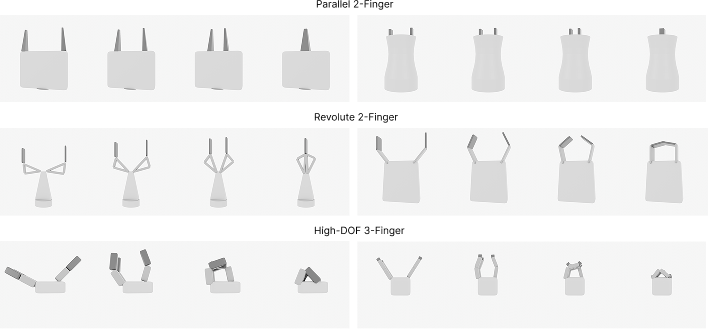}
  \caption{Examples of procedurally generated gripper of each category: parallel 2-finger, revolute 2-finger and high-dof 3-finger. Each row contains 2 randomly generated instances. We show 4 steps in gripper closing. The first one shows the state when the gripper is fully open and last one shows the state when the gripper is fully closed.}
  \label{fig:gripper_gen}
  \vskip -0.15in
\end{figure*}

\Figref{fig:gripper_gen} shows examples of our generated grippers together with its closing motion. In addition to the gripper's geometry, we also export the Swept Volume heuristic of the fully open to the half open state. The dimensions and translations from the base are automatically computed, given the configuration parameters of each gripper instance. We use COACD \cite{wei2022coacd} to compute the convex decomposition of each visual mesh for the collision mesh in Isaac-sim simulation.

\section{\model{} Dataset}
\label{appendix-dataset}
~~~Our cross-embodiment dataset used in our experiment features the largest simulation dataset for 6-DOF multi-embodiment grasping. We categorize the dataset into \model{}-Procedural and \model{}-Real to distinguish the grippers. In \Tableref{tab:related-datasets}, we summarize and compare with other 6-DOF grasping datasets. Ours \model{} dataset is 8X larger than the second largest dataset. For the generator training dataset, it consists of $2000 (\text{\# of sample grasps}) \times 25 (\text{\# of procedural grippers}) \times 3500 (\text{\# of training objects}) \approx 175M$ sampled grasps and grasp labels evaluated with the ACRONYM pipeline ~\cite{acronym2020} in Isaac-Sim simulation. Moreover, we also include a dataset for test objects of the grasps and labels sampled by $5000 (\text{\# of sample grasps}) \times 20 (\text{\# of real grippers}) \times 453 (\text{\# of test objects}) \approx 45M$. For the discriminator training dataset, we sampled on-generator grasps \cite{murali2025graspgen} with the generator and evaluated with the same labeling pipeline. The discriminator training dataset consists of another $2000 (\text{\# of sample grasps}) \times 25 (\text{\# of procedural grippers}) \times 3500 (\text{\# of training objects}) \approx 175M$ grasp samples and labels. In all, we have generated the largest multi-gripper 6-DOF grasping dataset. In total, we collect 350(=175+175)M grasps for training and our dataset contains a total of 395(=350+45)M grasps.

\begin{table*}[ht]
\centering
\resizebox{\linewidth}{!}{
    \begin{tabular}{@{}ccccccc@{}}
    \toprule
    Dataset          & Year                & \#Grippers      & \#Objects & \#Grasps        & Grasp Label   & Synthesis Method       \\ 
    \midrule
    HO-3D~\cite{hampali2020honnotate-ho3d} & 2020   & 1 (Human hand)             & 10                  & 78K                   & Only +ve               & Human Demo              \\

    EGAD~\cite{morrison2020egad}      & 2020        & 1 (2-finger)     & 2,331                & 233K                  & Only +ve               & Evolutionary Algorithm    \\

    DDG~\cite{liu2020deep-ddg}     & 2020           & 1 (5-finger)           & 500                 & 50K                   & Only +ve   & GraspIt + modified Q1 \\

    DexYCB~\cite{chao2021dexycb}       & 2021       & 1 (Human hand)             & 20                  & 582K                  & Only +ve               & Human Demo   \\
    
    Acronym ~\cite{acronym2020}  & 2021       & 1 (2-finger)     & 8,872                & 17.7M                 & +ve \& -ve & Flex~\cite{macklin2014unified-flex-simulator} \\
    UniGrasp ~\cite{shao2020unigrasp}  & 2020       & 12 (2 \& 3finger)     & 1000                &  2M+                   & Only +ve          & Contact Points Network + FastGrasp      \\
    
    DexGraspNet~\cite{wang2023dexgraspnet}  & 2023  & 1 (5-finger)           & 5,355                & 1.3M                  & Only +ve               & Differentiable grasping  \\
    Fast-Grasp'D~\cite{turpin2023fast-graspd} & 2023 & 3 (3-5 finger)       & 2,350                & 1M                    & Only +ve               & Differentiable grasping  \\
    MultiGripperGrasp \cite{Casas_MultiGripperGrasp_IROS2024}       & 2024               & 11 (2-5 finger \& Human) & 345        & 30.4M         & Ranked & GraspIt + Isaac Sim~\cite{nvidia2023-isaac-sim}  \\
    GraspGen \cite{murali2025graspgen}   & 2025           & 3 (2-finger \& Suction) & 8,515  & 53.1M         & +ve \& -ve & Sampling + Isaac Sim~\cite{nvidia2023-isaac-sim}  \\
    \midrule 
    GraspGen-X-Procedural & 2025 & 25 (3-finger, 2-finger procedural grippers) & 3500 & 350M & +ve \& -ve & Sampling + Isaac Sim~\cite{nvidia2023-isaac-sim} \\
    GraspGen-X-Real & 2025 & 20 (3-finger, 2-finger real-world grippers) & 453 & 45M & +ve \& -ve & Sampling + Isaac Sim~\cite{nvidia2023-isaac-sim} \\
    \bottomrule
\end{tabular}
}
\captionsetup{justification=centering}
\caption{Comparison of \model{} with existing grasping datasets. The table is adapted from \cite{murali2025graspgen}.}
\label{tab:related-datasets}
\vskip -0.2in
\end{table*}

\section{Simulation Experiments}
\label{appendix-simexps}
~~~In this section, we provide the details and additional results of simulation experiments.

\subsection{Experiment Setup}

\begin{figure*}[t]
  \centering
  \includegraphics[width=\linewidth]{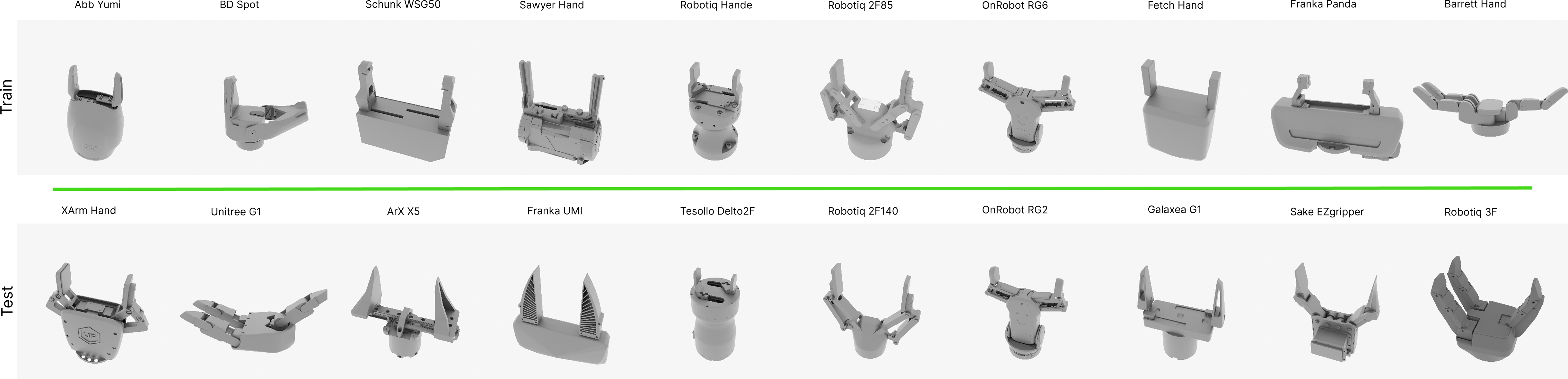}
  \caption{The set of training (upper row) and test (bottom row) grippers used in our experiments. We create a balanced split of all 20 grippers. }
  \label{fig:gripper_train_test}
  \vskip -0.2in
\end{figure*}

\begin{figure}[t]
  \centering
  \includegraphics[width=\linewidth]{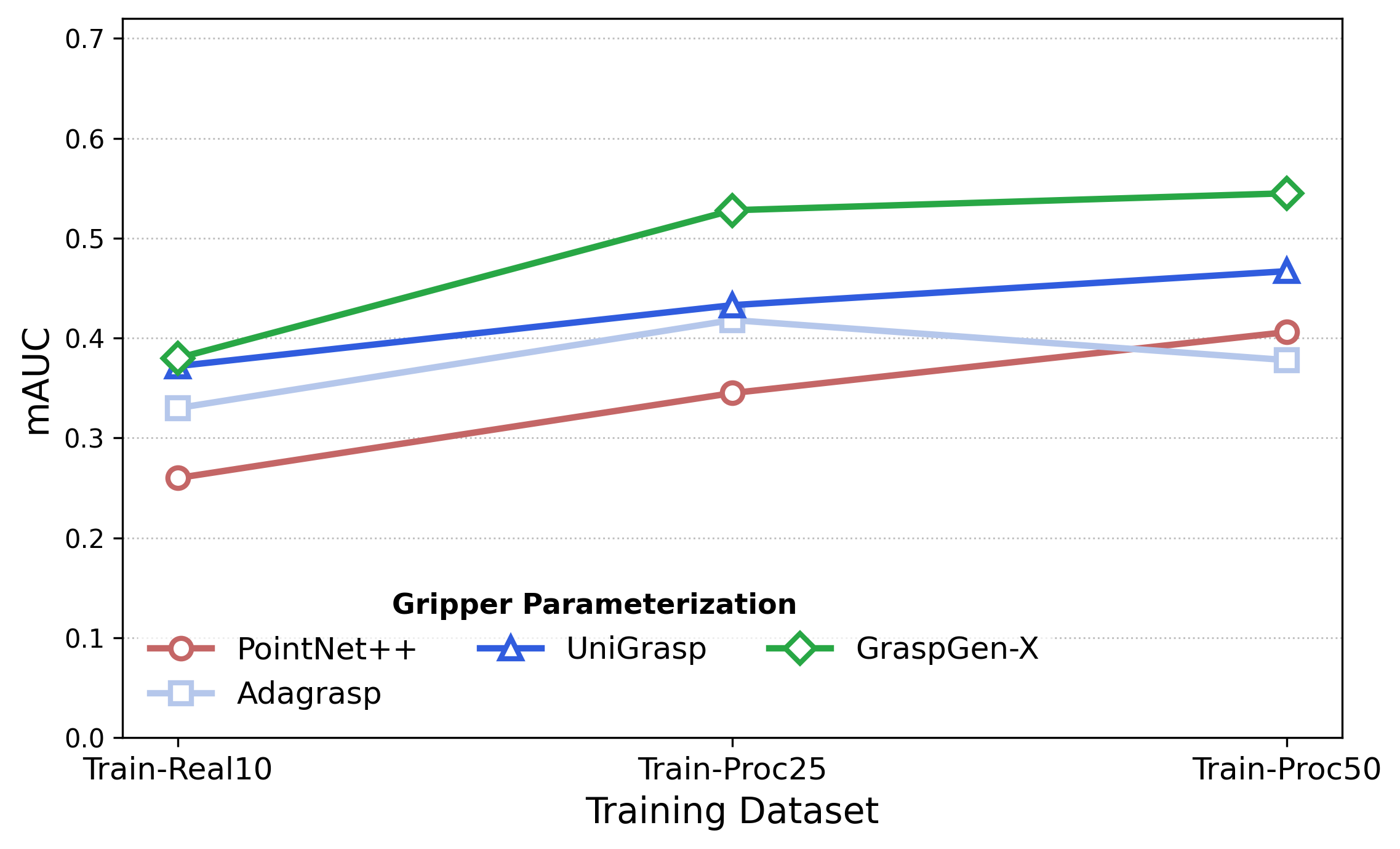}
  \caption{Results comparing \model{} and baselines' gripper encoding representations. We repeat the experiments in Sec \ref{sec:gripper_encoder} on different sets of grippers.}
  \label{fig:gripper_encoder_apx}
  \vskip -0.2in
\end{figure}

~~~\textbf{Train / Test Real Grippers.} \Figref{fig:gripper_train_test} shows the split of 20 grippers used in our experiments. We try to equally divide the set based on morphology similarity. We follow the same simulation setup for parallel grippers and 2-finger revolute grippers in ACRONYM \cite{acronym2020}. For 3-finger high-dof, we use PD controller to track a target joint trajectories. The target trajectory is a linear interpolation between the fully open state and fully closed state of each 3-finger gripper. 

\textbf{\model{} Training.} Our \model{} extends GraspGen with the additional 512-dim gripper embedding, concatenated with the 512-dim object embedding. We use the PointNet++~\cite{qi2017pointnet++} encoder for object embedding for training efficiency. For all other hyper-parameters, we follow the exact same setup in the GraspGen \cite{murali2025graspgen} codebase.

\subsection{Zero-shot Evaluation}

~~~We use mAUC, i.e., the average AUC of the precision-coverage curve of all test grippers and test objects, as our primary metric. For each gripper and each object, we plot the precision-coverage with the ranked 2k grasps iteratively.

\Tableref{tab:zero-shot-apx} shows the mAUC of each test gripper. We find that \model{} outperforms well in almost all grippers except OnRobot RG2, which falls significantly behind RTG. This suggests that there is still room for improvement in our end-to-end cross-embodiment model. We hypothesize that the OnRobot RG2's morphology may not be well covered with our procedural grippers, but this can be mitigated by further training on more procedural grippers. 

Nevertheless, our \model{} still achieves the SOTA performance across each category. \Figref{fig:grasp_vis_apx} visualizes the 5 generated grasps with \model{} for all test grippers.

 \begin{table}[ht]
      \centering
      \caption{Zero-shot performance on novel test grippers and novel test objects. We report the mAUC of each gripper.}
      \resizebox{0.9\linewidth}{!}{
      \begin{tabular}{cccc}
        \toprule
        & GraspGen-DTR & GraspGen-RTG & GraspGen-X \\
        \midrule
        ARX X5 & 0.270 & 0.532 & \textbf{0.620} \\
        Galaxea G1 & 0.319 & 0.537 & \textbf{0.663} \\
        Franka UMI & 0.267 & 0.256 & \textbf{0.441} \\
        Tesollo Delto2F & 0.002 & 0.134 & \textbf{0.285} \\
        \midrule
        Avg. Parallel 2F & 0.215 & 0.365 & \textbf{0.502} \\
        \midrule
        Sake EZgripper & 0.064 & 0.420 & \textbf{0.522} \\
        Robotiq 2F140 & 0.051 & 0.303 & \textbf{0.469} \\
        OnRobot RG2 & 0.002 & \textbf{0.241} & 0.136 \\
        XArm Hand & 0.015 & \textbf{0.551} & 0.525 \\
        \midrule
        Avg. Revolute 2F & 0.033 & 0.379 & \textbf{0.413} \\
        \midrule
        Unitree G1 & 0.269 & 0.662 & \textbf{0.818} \\
        Robotiq 3F & 0.002 & 0.343 &\textbf{0.579} \\
        \midrule
        Avg. High-DOF 3F & 0.136 & 0.503 & \textbf{0.699} \\
        \bottomrule
      \end{tabular}}
      \label{tab:zero-shot-apx}
      \vskip -0.2in
\end{table}

\subsection{Supervised Finetuning Adaptation}

~~~ We adopt the same metrics in generator training as those used in GraspGen \cite{murali2025graspgen}. For a given gripper embodiment $\mathcal{E}$ and object $\mathcal{O}$, we have a set of positive SE(3) grasp poses $G_{\mathcal{E}, \mathcal{O}}$ and the corresponding predictions from ~\model{} $\hat{G}_{\mathcal{E}, \mathcal{O}}$. For each grasp pose $\hat{g}\in\hat{G}_{\mathcal{E}, \mathcal{O}}$, we find the corresponding nearest neighbor grasp $g$ in the ground truth set $G_{\mathcal{E}, \mathcal{O}}$ based on the cost function defined in ~\cite{murali2025graspgen}. To capture both the accuracy of these grasp predictions, we separately compute the translation error (L2 distance in \textit{m}) and rotation error (the geodesic error in radians) between $\hat{g}$ and $g$, where $g = (R, t)$ and $\hat{g} = (\hat{R}, \hat{t})$. Similarly, to measure how well the generator captures the ground truth distribution, recall refers to the ratio of grasps in $G_{\mathcal{E}, \mathcal{O}}$ that have a nearest neighbour grasp in $\hat{G}_{\mathcal{E}, \mathcal{O}}$ within a threshold of $t_{dist}=$1\textit{cm}.




\textbf{Translation Error (m)}:
\begin{equation}
\mathcal{E}_{\mathrm{trans}} = \lVert t - \hat{t} \rVert_2
\end{equation}

\textbf{Rotation Error (radians)}:
\begin{equation}
\mathcal{E}_{\mathrm{rot}} = 
\arccos\!\left( \frac{\mathrm{tr}(R^\top \hat{R}) - 1}{2} \right)
\end{equation}

\textbf{Recall}:
\begin{equation}
\mathrm{Recall}
=
\frac{
\left|
\left\{
g \in G_{\mathcal{E}, \mathcal{O}}
\;\middle|\;
\exists\, \hat{g} \in \hat{G}_{\mathcal{E}, \mathcal{O}}
\text{ s.t. }
\lVert t - \hat{t} \rVert_2 < 0.01
\right\}
\right|
}{
\left| G_{\mathcal{E}, \mathcal{O}} \right|
}
\end{equation}

\subsection{Gripper Encoding Representation}

~~~We follow the experiment in \Figref{fig:ablation_study} to compare different gripper encoding representations used in previous work. The results show that \model{} gripper encoding performs better than baselines in different sets of training grippers. Moreover, training with 50 procedural grippers improves training with 25 procedural grippers also in the PointNet++~\cite{freiberg2024diffusionmultiembodimentgrasping} and AdaGrasp~\cite{xu2021adagrasp} (TSDF) representations.

\subsection{Ablation Study}

\begin{figure}[h]
    \centering 
    \begin{subfigure}{0.4\textwidth} 
        \includegraphics[width=\textwidth]{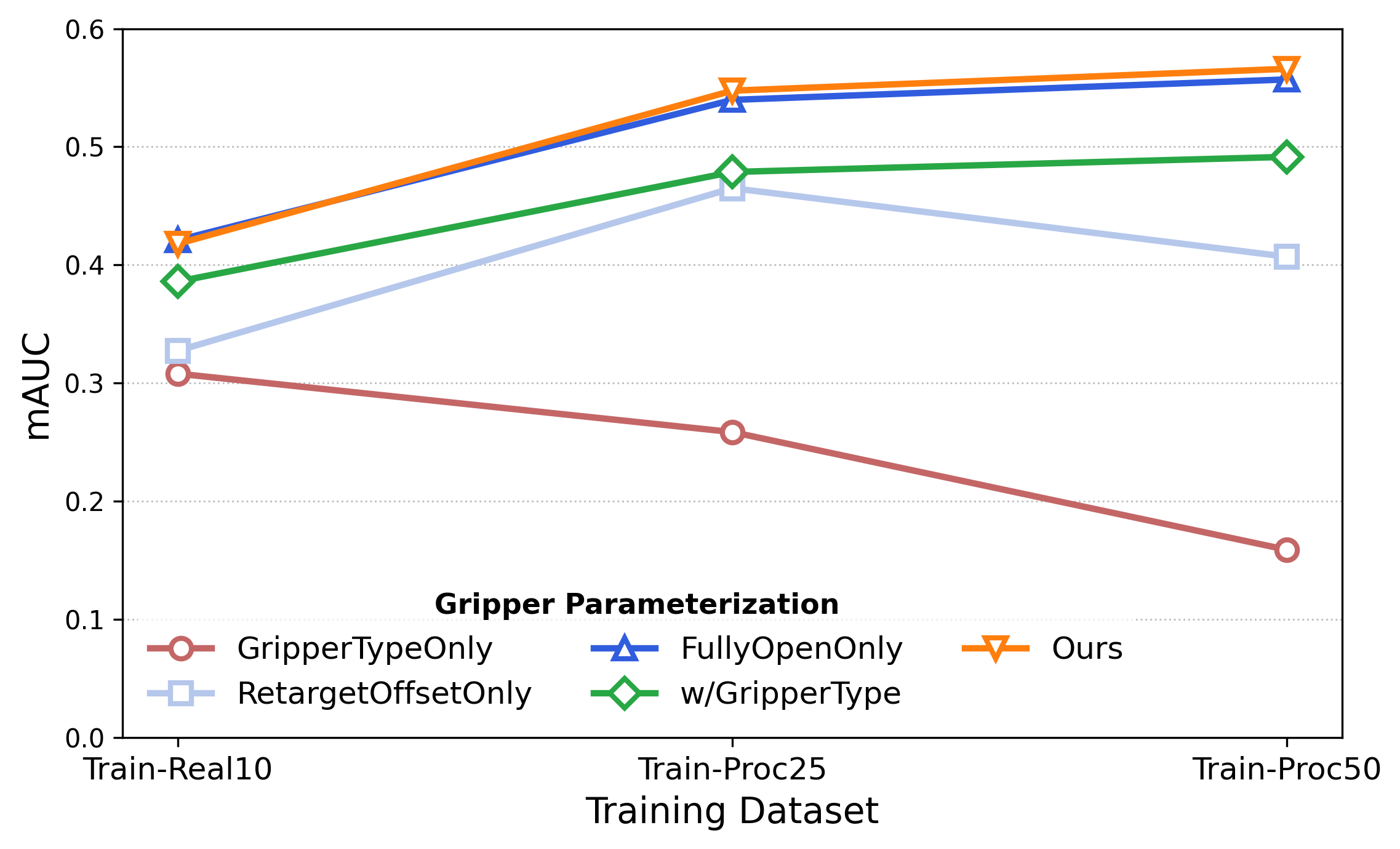} 
        \caption{Parallel 2-Finger}
        \label{fig:parallel2f_ablate}
    \end{subfigure}
    \begin{subfigure}{0.4\textwidth}
        \includegraphics[width=\textwidth]{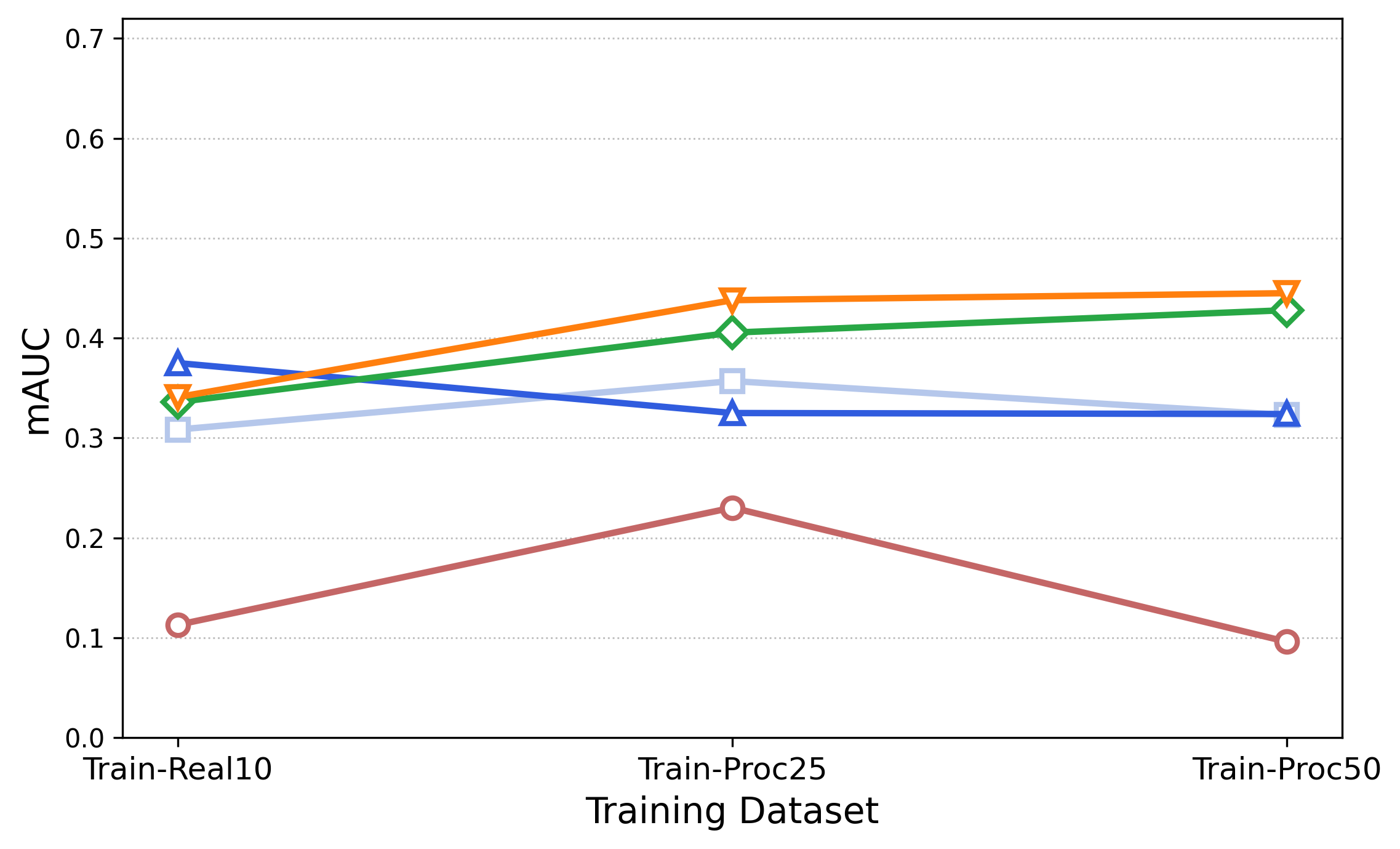}
        \caption{Revolute 2-Finger}
        \label{fig:revolute2f_ablate}
    \end{subfigure}
    \begin{subfigure}{0.4\textwidth}
        \includegraphics[width=\textwidth]{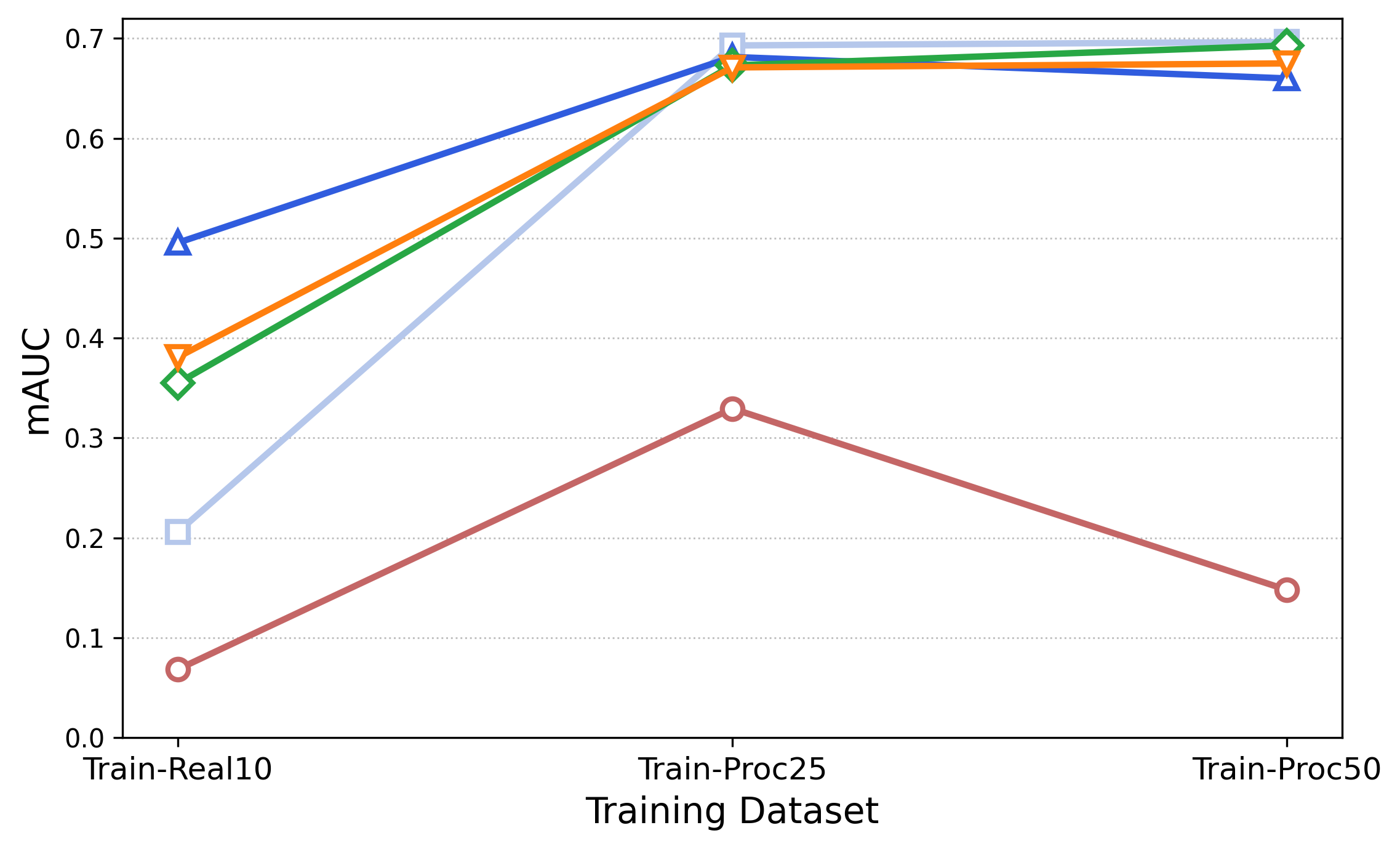}
        \caption{High-DOF 3-Finger}
        \label{fig:3-finger}
    \end{subfigure}
    \caption{Ablation on gripper parameterization and gripper training dataset. We show the results of each category: (a) parallel 2-finger gripper, (b) revolute 2-finger gripper, (c) high-dof 3-finger gripper.}
    \label{fig:ablation_per_category}
    \vskip -0.2in
\end{figure}

\begin{figure*}[t]
  \centering
  \includegraphics[width=\linewidth]{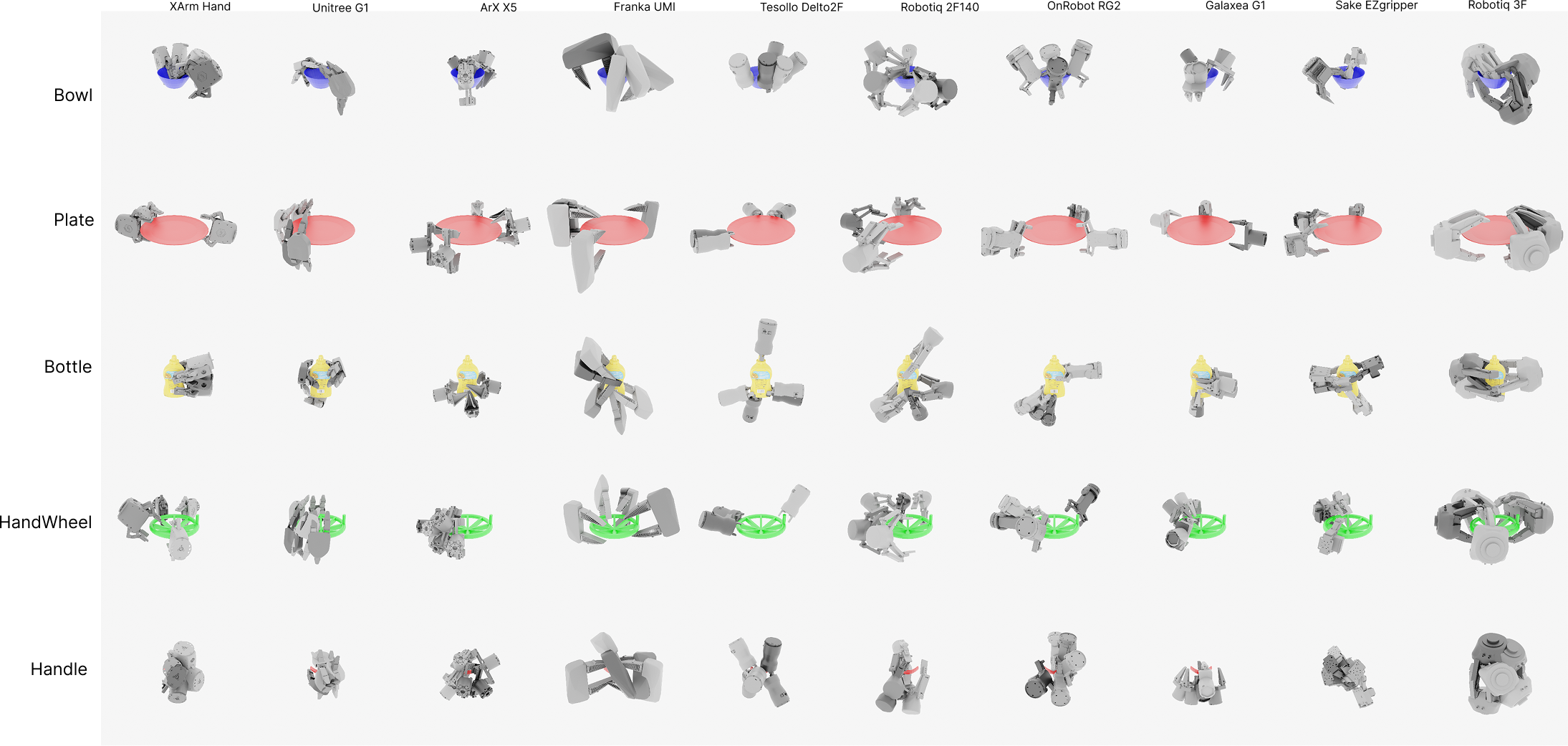}
  \caption{Visualization of \model{} generated grasps with on all 10 test grippers and on 5 novel objects.}
  \label{fig:grasp_vis_apx}
\end{figure*}

~~~ \Figref{fig:ablation_per_category} shows the results of ablation in the parameterization of the gripper and the distribution of the training grippers. We plot the results of average AUC over grippers of each category, including parallel 2-finger grippers, revolute 2-finger grippers, and high-dof 3-finger grippers.

Comparing FullyOpenOnly (6-dim) and ours (12-dim), we find that the gap primarily comes from the revolute 2-finger grippers. Many grippers in this category, e.g., Robotiq-2F140 and OnRobot-RG2, will rotate the fingers so that the fingertip will move forward along the z-axis during closing. Consequently, it is important to capture the information of the closing motion when computing the grasp pose. However, FullyOpenOnly does not contain this information in its parameterization while ours Swept Volume of fully open and half open states provides.

\section{Real Robot Experiment}
\label{appendix-realrobot}
~~~ \textbf{Industrial Manipulator:} Our UR10 setup consists of a single extrinsically calibrated RealSense D435 RGB-D camera overlooking the scene. We use SAM2~\cite{ravi2024sam2} running on a 6000 Ada GPU for segmentation, as well as FoundationStereo \cite{wen2025foundationstereo} for depth estimation. The input to \model{} is a partial pointcloud segmented on the target object. cuRobo~\cite{curobo_report23} and NVBlox~\cite{millane2024nvblox} running on a Jetson is used for collision-free motion planning. All models returned a set of predicted grasps and confidence scores. We use the top-100 grasps as pose targets for the motion planner, which filters out grasps that are in collision or do not have an IK solution.

\textbf{Low-Cost Arm:} We use AgileX Piper robot with its parallel 2-finger hand. We mount a ZED2 camera on the arm's end-effector. In this experiment, we assume access to the object model including a YCB Mustard bottle and a cube. We use FoundationPose \cite{wen2023foundationpose} to estimate an accurate 6D object pose. The input to \model{} is the complete pointcloud under the 6D pose. The target grasp pose is then transformed into the base frame of the robot. We then compute an IK solution in Pybullet and use a linearly interpolated trajectory to command the arm to the target joint configuration to execute the grasp.

\textbf{Humanoid Manipulator:} Additionally, we also validate \model{} on a Unitree G1 3-finger 7-DOF hand  and a stereo-camera mounted on the chest of G1 robot (\Figref{fig:coverfig}). Similarly to the low-cost arm experiment, we infer the grasp poses with the full object pointcloud and generate the grasping motion with a linearly interpolated trajectory from the initial joint configurations to the target configuration, which is computed via IK.

We test \model{} with the YCB mustard bottle with randomly placed stable poses. We observe that it achieves 100\% success rate out of 5 trials.

\section{Discussion}

\begin{figure}[ht]
  \centering
  \includegraphics[width=0.7\linewidth]{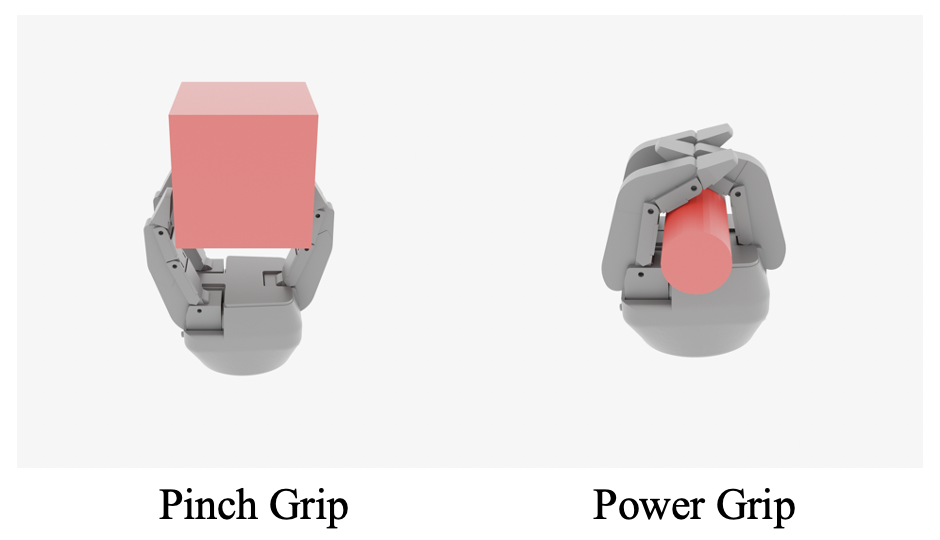}
  \caption{Visualization of sampled antipodal grasps of a 3-finger gripper. It can be either power grip or precision grip.}
  \label{fig:3-finger}
  \vskip -0.15in
\end{figure}

\begin{figure}[ht]
  \centering
  \includegraphics[width=0.7\linewidth]{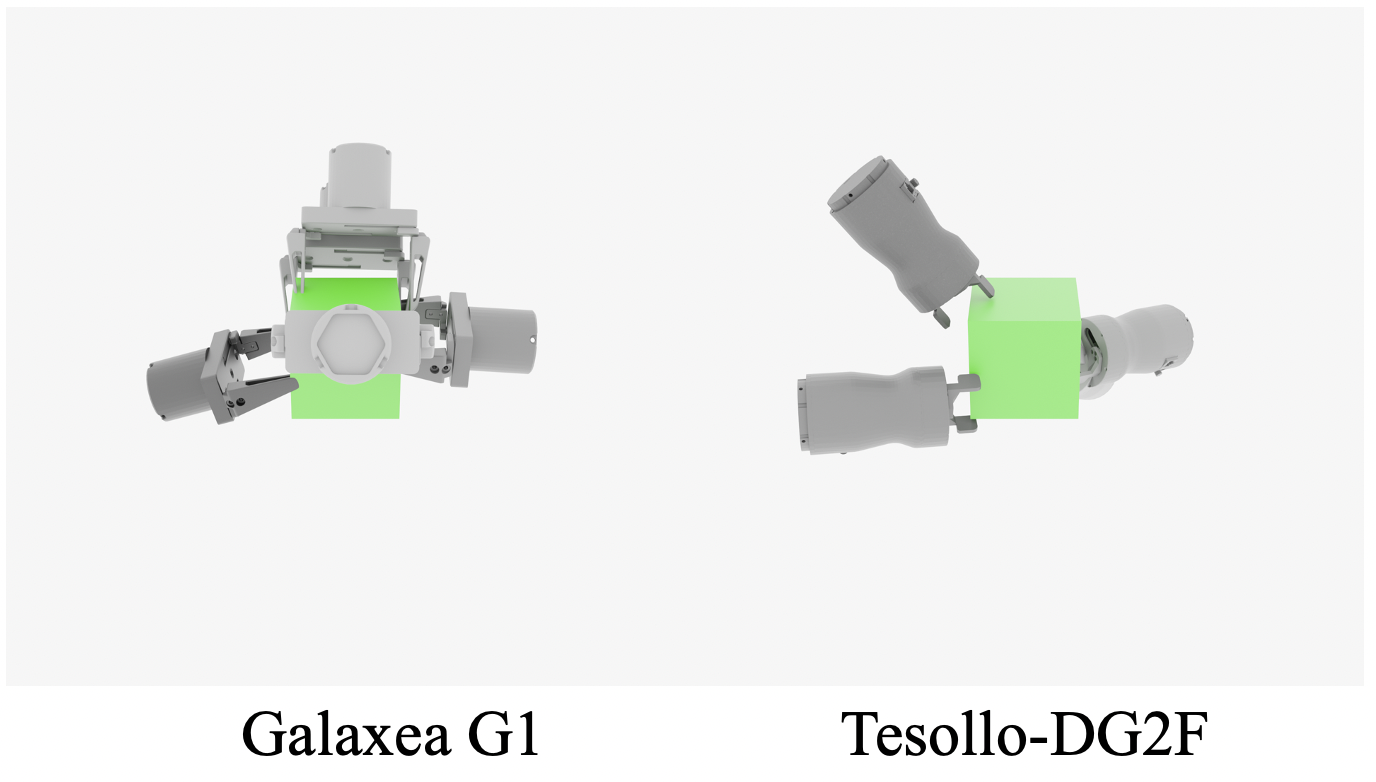}
  \caption{Visualization of \model{} generated grasps with on all 10 test grippers and on 5 novel objects.}
  \label{fig:non_grasp}
  \vskip -0.15in
\end{figure}

~~~\textbf{Non-graspable Objects.} \Figref{fig:non_grasp} visualizes the model's prediction of a non-graspable object. We choose a non graspable cube (12cm edge) with two Parallel045 2F grippers (Gallaxea G1
and Tesollo-DG2F) with a maximal opening width smaller than the cube size. The figure shows the top-ranked grasps of GraspGen-X. For all such grasps, the discriminator has a low confidence
score.

\textbf{Antipodal Sampling of 3-Finger Grippers.} 3-finger grippers are actuated to follow a predefined finger closure trajectory with PD control (\Figref{fig:3-finger}) to grip the object.  We assume that the thumb finger is stretched in the $+x$ direction and the other two fingers are stretched in the $-x$ direction. For antipodal sampling, we align $+x$ at the tip of the thumb with the normal of the contact point sampled on the object. Then, distance offset and pose orientation are randomized to search for positive grasps. We acknowledge that the distribution of end grasps is more limited in variation than the more general dexterous hand grasping problem, which is future work. 

\textbf{Sweep Volume Representation.} Our swept-volume representation omits some aspects of the gripper, such as its contact friction and kinematics. For example, in 3-finger grippers, it fails to reflect the asymmetric contact pattern of the $+x$ thumb finger and two $-x$ fingers, which may be crucial to grasp small objects. Despite the loss of information, the representation is an efficient representation for cross-embodiment 6-DOF grasping compared to prior methods and it works well on 3-finger and even 5-finger grippers.

\section{\model{} Post CVPR Update}

\subsection{Difference between CVPR and Latest Checkpoint}

Since the acceptance of the paper at CVPR 2026, we the trained the GraspGen-X model on a larger dataset to improve the zero-shot generalization performance across more grippers and objects.

\begin{table}[h]
    \centering
    \begin{tabular}{lcccc}
        \toprule
        Version & Procedural Grippers & Objects & Model & Total Grasps \\
        \midrule
        CVPR   & 25 & 3.5K & Gen + On-Generator Discriminator & 350M \\
        Latest  & 32 & 8.5K & Gen + On-Generator Discriminator & 2B \\
        \bottomrule
    \end{tabular}
    \caption{Comparison of the training datasets and models used in the CVPR and Latest versions of \model{}.}
    \label{tab:cvpr_vs_arxiv_dataset}
\end{table}

We will release the latest 2 Billion Grasp dataset soon.

\subsection{Grasp Mixture-of-Experts}
User feedback highlights a need for top-down grasps, particularly for benchmarks like LIBERO where objects rest in a upright stable poses. Although our diffusion model generates top-down grasps, it does not strictly enforce them. To address this in the code release, we introduce a Grasp Mixture-of-Experts (MoE) that samples from both the diffusion model and an Oriented Bounding Box (OBB). While the OBB sampler excels with simple cuboidal objects, the diffusion model handles complex, non-convex shapes and challenging poses. Finally, the \model{} discriminator scores all sampled grasps to ensure only high-quality outputs are returned.

\begin{algorithm}[h]
  \caption{Grasp Mixture-of-Experts (MoE) Sampling}
  \label{alg:grasp_moe}
  \begin{algorithmic}[1]
  \Require Object point cloud $\mathcal{O}$, gripper $\mathcal{R}$, target number of grasps $K$, \model{} generator $\pi_{\text{gen}}$, \model{} discriminator $\pi_{\text{dis}}$, OBB sampler $\pi_{\text{obb}}$
  \State Fit oriented bounding box: $\text{OBB} \gets \texttt{fitOBB}(\mathcal{O})$
  \State Sample OBB grasps: $\mathcal{G}_{\text{obb}} \gets \pi_{\text{obb}}(\text{OBB}, \mathcal{R})$
  \State Sample generator grasps: $\mathcal{G}_{\text{gen}} \gets \pi_{\text{gen}}(\mathcal{O}, \mathcal{R})$
  \State Aggregate expert grasps: $\mathcal{G}_{\text{moe}} \gets \mathcal{G}_{\text{obb}} \cup \mathcal{G}_{\text{gen}}$
  \State Score with discriminator: $\mathcal{S}_{\text{moe}} \gets \pi_{\text{dis}}(\mathcal{G}_{\text{moe}}, \mathcal{O}, \mathcal{R})$
  \State Filter top-$K$ grasps: $\mathcal{G}_{\text{top-}K} \gets \texttt{rank}(\mathcal{G}_{\text{moe}}, \mathcal{S}_{\text{moe}}, K)$
  \State \Return $\mathcal{G}_{\text{top-}K}$
  \end{algorithmic}
\end{algorithm}

In our repository, the Grasp Mixture-of-Experts (MoE) is enabled by default. The OBB sampler $\pi_{\text{obb}}$ uses Principal Component Analysis (PCA) on the object point cloud to fit an OBB along its longest edges, assuming the positive z-axis aligns with gravity. It samples both top-down and side grasps. To ensure physical realism, grasps are excluded from any OBB face that exceeds the robot gripper's stroke length. Ultimately, this MoE framework combines the strengths of both generative experts, relying on our discriminator's robustness to filter out poor-quality grasps.

\end{document}